\documentclass[11pt]{article}

\usepackage[final]{acl}

\usepackage{times}
\usepackage{latexsym}

\usepackage[T1]{fontenc}

\usepackage[utf8]{inputenc}

\usepackage{microtype}

\usepackage{inconsolata}

\usepackage{graphicx}

\usepackage{amsmath}
\usepackage{amssymb}
\usepackage[utf8]{inputenc}
\usepackage{geometry}
\usepackage{booktabs}   
\usepackage{multirow}
\usepackage{xcolor}     
\usepackage{colortbl}   
\usepackage{array}
\usepackage{subcaption}
\usepackage{mathtools}
\usepackage{dsfont}
\usepackage{xurl}

\definecolor{HighlightRow}{RGB}{235, 242, 250} 
\definecolor{ModelGray}{RGB}{242, 242, 242}
\definecolor{GreenText}{RGB}{10, 10, 245}
\definecolor{RedText}{RGB}{214, 39, 40}         

\newcommand{\valup}[2]{#1\textsubscript{\textcolor{GreenText}{\textbf{+#2}}}}
\newcommand{\valdown}[2]{#1\textsubscript{\textcolor{GreenText}{\textbf{#2}}}}

\newcommand{\accup}[2]{#1\textsubscript{\textcolor{GreenText}{\textbf{+#2}}}}
\newcommand{\accdown}[2]{#1\textsubscript{\textcolor{RedText}{\textbf{-#2}}}}

\newcommand{\tokdown}[2]{#1\textsubscript{\textcolor{GreenText}{\textbf{-#2}}}}
\newcommand{\tokup}[2]{#1\textsubscript{\textcolor{RedText}{\textbf{+#2}}}}

\title{SHAPE: Stage-aware Hierarchical Advantage via Potential Estimation for LLM Reasoning}

\author{
  Zhengyang Ai\textnormal{\textsuperscript{1}},
  Zikang Shan\textnormal{\textsuperscript{2}},
  Xiaodong Ai\textnormal{\textsuperscript{1}},
  Jingxian Tang\textnormal{\textsuperscript{1}},
  Hangkai Hu\textnormal{\textsuperscript{1}},
  Pinyan Lu\textnormal{\textsuperscript{1,3}}\thanks{Corresponding author.} \\
  $^{\text{1}}$Huawei Taylor Lab \\
  $^{\text{2}}$Center for Data Science, Peking University \\
  $^{\text{3}}$Shanghai University of Finance and Economics \\
  \small \texttt{aizhengyang@huawei.com}, \texttt{lu.pinyan@mail.shufe.edu.cn}
}

\begin{document}
\maketitle
\begin{abstract}

  Process supervision has emerged as a promising approach for enhancing LLM reasoning, yet existing methods fail to distinguish meaningful progress from mere verbosity, leading to limited reasoning capabilities and unresolved token inefficiency. To address this, we propose Stage-aware Hierarchical Advantage via Potential Estimation (SHAPE), a framework that formalizes reasoning as a trajectory through a state space of empirical solvability. SHAPE introduces a hierarchical credit assignment mechanism: at the \textit{segment level}, it employs a stage-aware advantage function to prioritize efficient breakthroughs in low-potential states; at the \textit{token level}, it utilizes entropy-driven redistribution to sharpen execution signals. Extensive experiments in math reasoning across three base models and five benchmarks demonstrate that SHAPE achieves an average accuracy gain of 3\% with 30\% reduced token consumption.
  
\end{abstract}

\section{Introduction}
\label{intro}

Reinforcement Learning (RL) has emerged as the standard paradigm for post-training Large Language Models (LLMs). While outcome-based methods such as Group Relative Policy Optimization (GRPO) \citep{shao2024deepseekmath} optimize against final answer correctness, they fundamentally struggle with sparse rewards, often leading to inefficient exploration or overthinking due to reward misspecification. Although learned Process Reward Models (PRMs) \citep{lightman2023let, zhang2025lessons, cui2025process} offer dense feedback to mitigate this, they incur prohibitive annotation costs and remain vulnerable to reward hacking. Consequently, the field has increasingly shifted toward \textit{rule-based process supervision} \citep{qu2025optimizing, guo2025segment, nie2026attnpo}, which derives dense signals via rule-based estimation rather than unreliable black-box verifiers, facilitating explicit designs to robustly mitigate reward hacking.

The core of this paradigm lies in estimating the value of intermediate states (i.e., segments) to fill the void of sparse outcome signals. In this work, we unify this intermediate value under the concept of \textit{Reasoning Potential ($\Phi$)}. In classical RL, potential serves as a scalar function representing the latent value of a state—acting not as a direct environmental reward, but as a form of preference or prior knowledge regarding state quality \citep{ng1999policy,wiewiora2003potential}. In LLM reasoning, $\Phi$ acts as a dynamic progress gauge: a low $\Phi$ reflects a state of high uncertainty or confusion, while a rising $\Phi$ signifies that the reasoning path is effectively bridging the logical gap to the solution.

\begin{figure}[t]
  \centering
  \includegraphics[width=\linewidth]{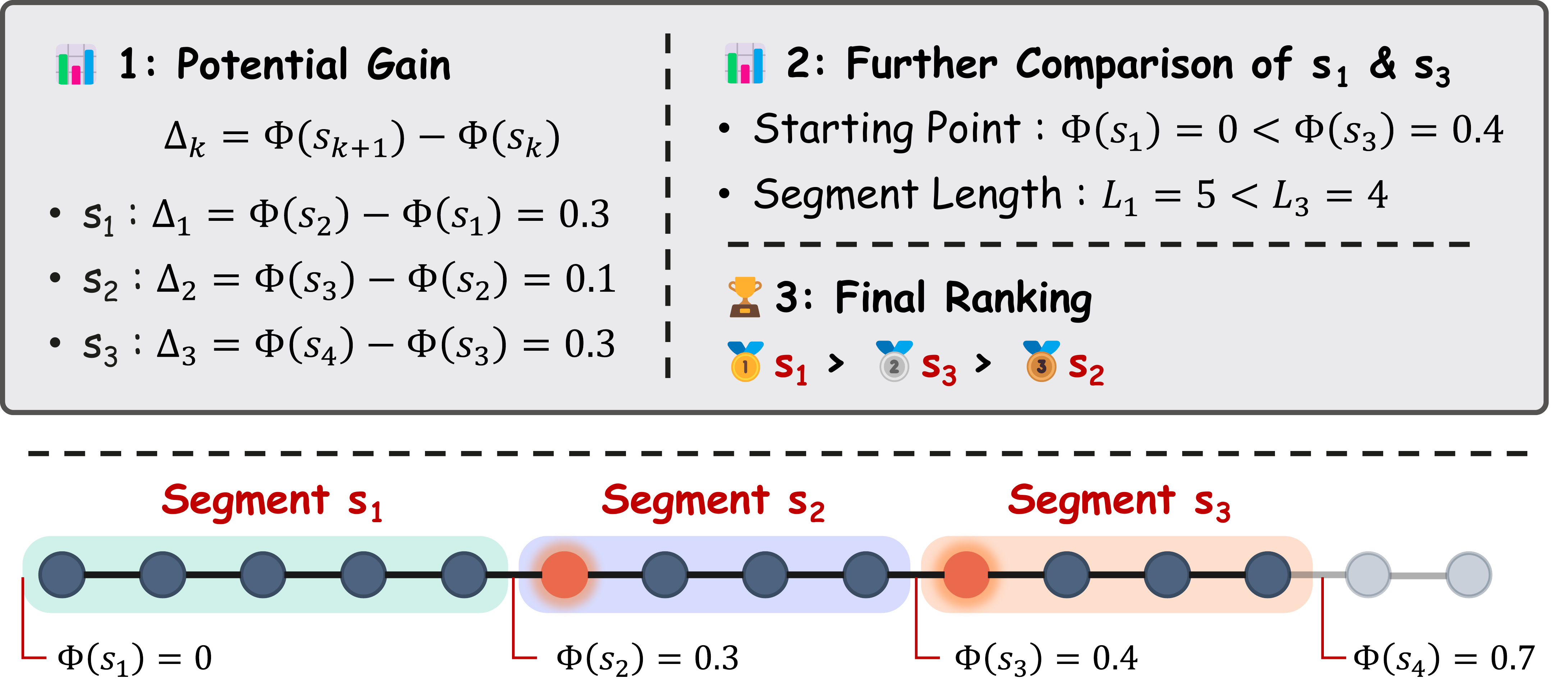}
  \caption{{Illustration of optimal reasoning path.} Segment $s_2$ ranks lowest due to insufficient Potential Gain ($\Delta=0.1$). While $s_1$ and $s_3$ share identical gain ($\Delta=0.3$), $s_1$ achieves the highest rank by jointly satisfying {Stage Awareness} (breaking through from a harder state $\Phi=0$) and {Efficiency} (shorter length $L_1 < L_3$).}  \label{fig:p1}
\end{figure}

With the concept of potential established, the critical question becomes: \textit{How can we leverage $\Phi$ to construct dense process rewards for effective reward shaping?} We argue that an optimal reasoning path (i.e., a segment transition) must jointly satisfy three essential criteria (as illustrated in Figure~\ref{fig:p1}):

\begin{enumerate}
  \item \textbf{Potential Gain:} The fundamental requirement is that a segment must effectively bridge the logical gap to the solution ($\Delta \Phi > 0$).
  
  \item \textbf{Stage Awareness:} Not all gains are equal. Breakthroughs from \textit{lower-potential stages} (confusing foothills) represent overcoming high uncertainty and are thus more valuable than marginal refinements in \textit{high-potential stages} (near the summit).
  
  \item \textbf{Token Efficiency:} Among paths with equal gain, shorter paths should be prioritized. Verbose reasoning chains that accumulate computational cost without proportional progress must be discouraged.
\end{enumerate}

Existing approaches, however, fail to unify these dimensions. SPO \citep{guo2025segment} focuses solely on {Potential Gain}, ignoring stage difficulty. MRT \citep{qu2025optimizing} implicitly captures {Stage Awareness} but lacks checks for stepwise progress and neglects {Efficiency}. S-GRPO \citep{dai2025sgrpo} addresses length penalties but lacks the semantic granularity of potential guidance. Consequently, a unified framework optimizing all three dimensions remains absent.

We propose \textbf{SHAPE} (\textbf{\underline{S}}tage-aware \textbf{\underline{H}}ierarchical \textbf{\underline{A}}dvantage via \textbf{\underline{P}}otential \textbf{\underline{E}}stimation), a framework explicitly designed to satisfy these three principles simultaneously. Grounded in Potential-Based Reward Shaping (PBRS) \citep{ng1999policy,wiewiora2003potential}, SHAPE introduces a novel advantage function controlled by a {dynamic, length-dependent discount factor ($\gamma_k$)}. This single mechanism naturally encodes the optimal criteria: it measures \textit{Potential Gain} via difference modeling, enforces \textit{Stage Awareness} by scaling the baseline penalty according to the current stage's difficulty, and regulates \textit{Token Efficiency} by dynamically adjusting the discount rate based on segment length.

To further refine this guidance, SHAPE operates in a {hierarchical} manner. We introduce an entropy-driven {token-level redistribution} mechanism to implement fine-grained shaping atop the segment-level foundation. This sharpens the feedback by assigning higher credit to pivotal tokens, ensuring that the learning signal focuses on critical decision points within the segment.

Our contributions are summarized as follows:

\begin{itemize}
    \item  We formalize three essential 
          principles for optimal reasoning paths, and show how existing methods 
          only partially satisfy them.
    \item We propose SHAPE, which utilizes a dynamic discounting mechanism to jointly satisfy these criteria, supplemented by hierarchical token-level refinement.
    
    \item We demonstrate consistent  empirical gains across various benchmarks with significantly reduced token costs, validating the importance of stage-aware hierarchical reward shaping.
\end{itemize}

\section{Related Work}

\paragraph{RL for LLM Reasoning}
Reinforcement learning has been proven effective in improving LLM reasoning capabilities~\citep{jaech2024openai,guo2025deepseek}. The Reinforcement Learning with Verifiable Rewards (RLVR) paradigm trains the pretrained LLM in reasoning-heavy tasks~\citep{shao2024deepseekmath,guo2024deepseek} via large-scale RL on rule-based verifiers. The prominant algorithm, GRPO~\cite{shao2024deepseekmath}, replaces the value model of PPO~\citep{schulman2017proximal} with sample-based baseline, trading fine-grained feedback for reduced training resources. Despite later algorithmic improvements~\cite{liu2025understanding,yu2025dapo}, RLVR suffers from the sparsity of the outcome-based reward signal, leading to issues like training instability and reward misspecification (e.g. overthinking~\citep{chen2024not,zhang2025s1bench} and underthinking~\citep{qu2025optimizing}).

\paragraph{Fine-grained Credit Assignments}
Methods that provide fine-grained feedbacks have been proposed. \citet{yuan2025s,zhu2025vrpo} revisit PPO to improve value modeling, while \citet{kazemnejad2024vineppo,guo2025segment} estimate values with online rollouts. Such methods involve extra cost of either training a value model or doing expensive rollouts. Process reward models~\citep{lightman2023let}, either explicitly~\citep{zhang2025lessons} or implicitly~\cite{zhong2024dpo,cui2025process}, are leveraged to provide extra fine-grained feedback. These methods require pretraining a reward model and are prone to reward hacking. There are also works that uses reward bonus. ATTNPO~\citep{nie2026attnpo} further explores low-overhead step-level supervision by leveraging intrinsic attention signals to reduce redundant reasoning. Notably, MRT~\cite{qu2025optimizing} introduces a rule-based progress reward that is not hackable and efficient to compute, leading to higher token efficiency. Compared to MRT, our proposed framework takes into account more design principles and results in even better performance and token efficiency.

\section{Preliminaries: MRT}
\label{subsec:mrt}

To address sparse rewards, Meta Reinforcement Fine-Tuning (MRT) 
\citep{qu2025optimizing} introduces dense intermediate feedback. We 
summarize its core mechanism:

\paragraph{1. Trajectory Segmentation.}
The reasoning trajectory $y$ is decomposed into segments 
$S = (s_1, \ldots, s_K)$ using delimiters (e.g., newlines), with each 
boundary denoted as $s_k$.

\paragraph{2. Potential Estimation.}

At each boundary $s_k$, MRT generates $m$ rollouts, where each rollout receives a binary score $r_i \in \{0, 1\}$ based on correctness. The potential is simply the average score:
\begin{equation}
\Phi(s_k) = \frac{1}{m}\sum_{i=1}^{m} r_i .
\label{eq:prelim_mrt_phi}
\end{equation}
We term $\Phi(s_k)$ the {reasoning potential}—it quantifies the immediate solvability of the state $s_k$.

\paragraph{3. Advantage Computation.}
MRT defines a progress bonus based on the gap between the final outcome $R_{\text{outcome}}$ and the current potential. The advantage for segment $s_k$ is:
\begin{equation}
A_k^{\text{MRT}} = R_{\text{outcome}} + \alpha \cdot (R_{\text{outcome}} - \Phi(s_k)),
\label{eq:mrt_advantage}
\end{equation}
where $\alpha$ balances the outcome and progress signals. Notably, this formulation creates a {telescoping effect}: when summed over a trajectory, intermediate potentials cancel out, effectively reducing the objective to optimizing the final outcome relative to the initial baseline.

\paragraph{Key Insight.}
This design creates a {global drive toward correctness}: segments farther from the answer (lower $\Phi$) receive stronger bonuses upon success. This naturally encodes {stage-aware} incentives, rewarding breakthroughs from low-potential states more than marginal refinements near the solution.

\paragraph{Limitations.}
Despite its merits, MRT's formulation exhibits critical structural flaws:

\begin{enumerate}

    \item \textbf{Weak Local Incentives (Sandbagging Risk):} 
    The advantage function relies solely on the potential of the \textit{current} state relative to the endpoint ($R - \Phi(s_k)$), ignoring the potential change to the \textit{next} state ($\Phi(s_{k+1})$). This decoupling fails to enforce monotonic progress. Consequently, the model is not penalized for actions that decrease potential; instead, it can be rewarded for recovering from self-inflicted low-potential states. This structural loophole incentivizes {strategic sandbagging}—deliberately traversing circuitous, low-potential paths to maximize cumulative bonuses through repeated "recoveries", leading to reasoning degradation. We empirically analyze this phenomenon in \S~\ref{sandbagging}.
    
    \item \textbf{Efficiency Neglect:} 
    MRT lacks explicit length penalties. Verbose segments achieving the same potential gain as concise ones receive identical rewards, inadvertently encouraging computational inefficiency.
    
    \item \textbf{Uniform Token Credit:} 
    The segment-level advantage $A_k$ is broadcast uniformly to all tokens in $s_k$, failing to distinguish critical decision steps from trivial tokens.
\end{enumerate}

These limitations motivate SHAPE: we introduce {local} 
potential-difference modeling with length-dependent discounting to enforce 
monotonic progress while penalizing verbosity, and refine token-level 
credit redistribution.

\section{Method: The SHAPE Framework}

\begin{figure*}[h]
	\centering
	\centerline{\includegraphics[width=\textwidth]{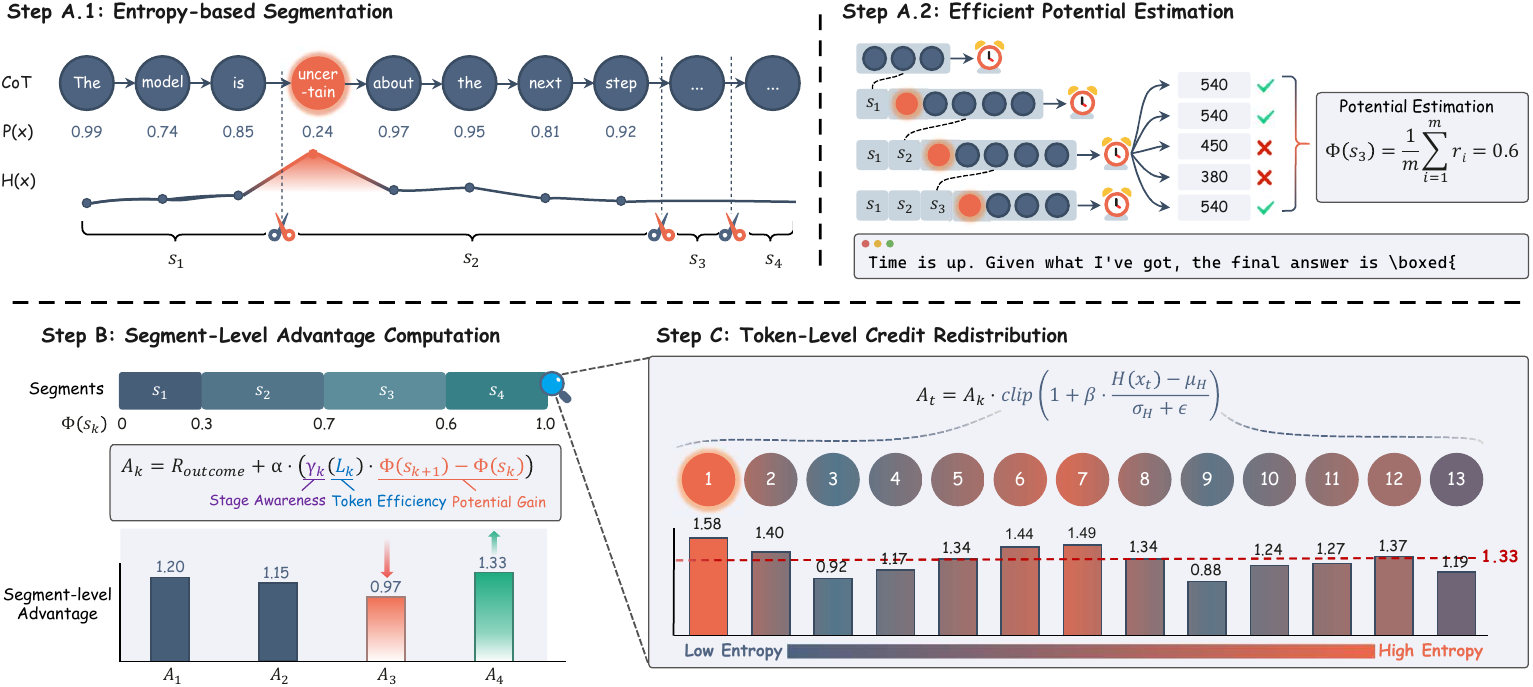}}
    \caption{{Overview of the SHAPE framework.} The pipeline consists of three steps: (A) decomposing reasoning to estimate state potentials (\S\ref{sec:path_construction}); (B) computing stage-aware segment-level advantages (\S\ref{sec:segment_advantage}); and (C) redistributing credit to sharpen token-level learning signals (\S\ref{sec:token_redistribution}).}
\label{ov}
\end{figure*}

In this section, we propose \textbf{SHAPE} (\textbf{\underline{S}}tage-aware \textbf{\underline{H}}ierarchical \textbf{\underline{A}}dvantage via \textbf{\underline{P}}otential \textbf{\underline{E}}stimation) framework. The overall architecture and workflow are illustrated in Figure~\ref{ov}.

\subsection{Trajectory Segmentation and Potential Estimation}
\label{sec:path_construction}

Standard process supervision typically relies on rigid delimiters. To capture true semantic transitions, we adopt the {Adaptive Cutpoint-based Partition} strategy from SPO \citep{guo2025segment}. However, while SPO relies on low-probability tokens to identify cutpoints, we argue that {high information uncertainty} is a more robust indicator of logical branching \citep{wang2025beyond, cheng2025reasoning}. Therefore, we adapt their framework to utilize token-level entropy as the segmentation criterion.

\paragraph{Entropy-Based Segmentation.}
We identify reasoning boundaries at points of high token-level entropy, which signal pivotal logical transitions:
\begin{equation}
    H(x_t) = -\sum_{v \in \mathcal{V}} \pi_\theta(v \mid x_{<t}) \log \pi_\theta(v \mid x_{<t}).
\end{equation}
Positions exceeding a threshold $\tau$ serve as candidate cutpoints. Following SPO, we downsample these candidates to standardize the trajectory into $K$ segments $S = (s_1, \dots, s_K)$, ensuring boundaries align with semantic decision nodes.

\paragraph{Efficient Potential Estimation.}
At each boundary $s_k$, we estimate the potential $\Phi(s_k)$ using the rollout formulation defined in Equation~\eqref{eq:prelim_mrt_phi}. Specifically, we execute $m$ forced-termination rollouts to calculate the expected success rate.
To maintain computational feasibility, we leverage \texttt{vLLM}'s {Prefix Caching} to avoid re-computing shared contexts and strictly limit the rollout length (e.g., \texttt{max\_tokens=16}), ensuring that the overhead of this spliced reasoning evaluation remains within an acceptable range. Detailed analysis of the trade-off between segmentation granularity and computational overhead is provided in \S~\ref{sec:granularity_impact}.

\subsection{Segment-Level Advantage Computation}
\label{sec:segment_advantage}

\paragraph{MDP Formulation and Standard PBRS.}

We formulate the reasoning process as a segment-level Markov Decision Process (MDP) to rigorously ground stepwise credit assignment in reinforcement learning theory. In this framework, states correspond to segment boundaries $s_k$, and transitions represent generating the next segment. To address sparse outcome supervision, we adopt Potential-Based Reward Shaping (PBRS) \citep{ng1999policy, wiewiora2003potential}:
\begin{equation}
F(s_k, s_{k+1}) = \gamma \Phi(s_{k+1}) - \Phi(s_k),
\label{eq:standard_pbrs}
\end{equation}
where $\gamma \in (0, 1]$ is a constant discount factor. This form guarantees 
\textit{policy invariance}: adding $F$ to the reward accelerates 
learning without altering the optimal policy.

\paragraph{Efficiency-Aware Modification.}
However, in the context of CoT reasoning, strictly preserving the original policy is suboptimal. The original objective typically ignores computational cost, potentially encouraging the model to generate verbose or redundant reasoning paths to maximize confidence.
To introduce an inductive bias favoring conciseness, we modify the discount factor to be a dynamic coefficient $\gamma_k$ dependent on the segment length $L_k$. We formalize this using a linear decay function:
\begin{equation}
    \gamma_k(L_k) = \max\left(\gamma_{\min}, \ 1 - \frac{L_k}{L_{\text{ref}}} (1 - \gamma_{\min})\right),
    \label{eq:dynamic_gamma}
\end{equation}
where $\gamma_{\text{min}} < 1$ represents the lower bound of the discount, and $L_{\text{ref}}$ is a manually configured reference length based on the expected granularity of reasoning steps. This formulation establishes a negative correlation between segment length and the discount factor: as $L_k$ increases, $\gamma_k$ decays linearly from $1$ down to the lower bound $\gamma_{\min}$. In our experiments, we empirically set the floor $\gamma_{\min}=0.9$; for a detailed ablation study on this hyperparameter, please refer to \S~\ref{sec:ablation}.

\paragraph{Advantage Definition.}
Incorporating this dynamic factor, we define the advantage $A_k$ for segment $s_k$ as:
\begin{equation}
    A_k = R_{\text{outcome}} + \alpha \cdot \underbrace{\left( \gamma_k(L_k) \cdot \Phi(s_{k+1}) - \Phi(s_k) \right)}_{\text{Potential-Based Shaping}}.
    \label{eq:final_advantage}
\end{equation}
Although introducing a variable discount factor theoretically alters the original policy, we prove in Appendix~\ref{subsec:task_consistency} that this formulation preserves \textit{Task Consistency}: correct solutions, regardless of their length, consistently yield higher total rewards than incorrect ones, preventing the model from exploiting the length penalty to generate short but wrong answers.

\paragraph{Mechanism Analysis.}
To elucidate how this formulation achieves our dual objectives—stage awareness and token efficiency—we decompose the shaping term. Let $\Delta_k = \Phi(s_{k+1}) - \Phi(s_k)$ be the raw potential gain. Substituting $\Phi(s_{k+1}) = \Phi(s_k) + \Delta_k$ into the shaping term (ignoring $\alpha$ for analysis), we obtain:
\begin{equation}
    F_k \approx \Delta_k - \underbrace{(1 - \gamma_k(L_k)) \cdot \Phi(s_k)}_{\text{Reasoning Tax}}.
    \label{eq:tax_decomposition}
\end{equation}
This decomposition reveals that the effective reward is the raw gain $\Delta_k$ minus a "Tax". This Tax term naturally enforces our two design goals:
\begin{itemize}
    \item \textbf{Stage Awareness:} The tax is proportional to the baseline potential $\Phi(s_k)$. In early reasoning stages where confidence is low ($\Phi(s_k)$ is small), the tax is negligible, encouraging the model to attempt breakthroughs. Conversely, in high-confidence states, the tax increases, suppressing potential inflation.
    \item \textbf{Token Efficiency:} The tax is proportional to $(1 - \gamma_k)$, which grows linearly with segment length $L_k$ (as defined in Equation~\eqref{eq:dynamic_gamma}). Longer segments incur a heavier tax, compelling the model to justify extra tokens with substantial potential gains.
\end{itemize}
We provide the mathematical derivation of these properties in Appendix~\ref{app:properties}.

\begin{table*}[h!]
  \centering

  \setlength{\tabcolsep}{1.5pt} 
  
  \resizebox{\textwidth}{!}{%
      \scriptsize 
      \begin{tabular}{l >{\hspace{4mm}}l llll lllll | ll}
      \toprule
       & 
      \multicolumn{2}{c}{\textbf{AIME 24}} & 
      \multicolumn{2}{c}{\textbf{AIME 25}} & 
      \multicolumn{2}{c}{\textbf{AMC 23}} & 
      \multicolumn{2}{c}{\textbf{MATH500}} & 
      \multicolumn{2}{c}{\textbf{Minerva}} & 
      \multicolumn{2}{c}{\textbf{Overall}} \\ 
      \cmidrule(lr){2-3} \cmidrule(lr){4-5} \cmidrule(lr){6-7} \cmidrule(lr){8-9} \cmidrule(lr){10-11} \cmidrule(lr){12-13}
      \textbf{Method} & Acc & Tokens & Acc & Tokens & Acc & Tokens & Acc & Tokens & Acc & Tokens & Acc & Tokens \\
      \midrule
      
      \rowcolor{ModelGray}
      \multicolumn{13}{l}{\textbf{\textit{DeepSeek-R1-Distill-Qwen-1.5B}}} \\
      GRPO            & 34.7 & 8772 & 27.5 & 8109 & 79.1 & 5091 & 84.8 & 3354 & 34.2 & 5228 & 52.1 & 6111 \\
      MRT             & 33.1 & 6577 & 28.6 & 6085 & 79.3 & 4058 & 85.0 & 2734 & 33.6 & 3705 & 51.9 & 4632 \\
      \rowcolor{HighlightRow}
      SHAPE  & \valup{37.1}{2.4} & \valdown{6164}{-29.7\%} 
                & \valup{31.8}{4.3} & \valdown{5425}{-33.1\%} 
                & \valup{81.5}{2.4} & \valdown{3612}{-29.1\%} 
                & \valup{87.8}{3.0} & \valdown{2415}{-28.0\%} 
                & \valup{35.5}{1.3} & \valdown{3207}{-38.7\%} 
                & \valup{54.7}{2.6} & \valdown{4165}{-31.8\%} \\
      \midrule
      
      \rowcolor{ModelGray}
      \multicolumn{13}{l}{\textbf{\textit{DeepScaleR-1.5B-Preview}}} \\
      GRPO            & 38.6 & 7106 & 35.7 & 7041 & 82.1 & 4797 & 87.1 & 3169 & 34.5 & 4965 & 55.6 & 5416 \\
      MRT             & 41.3 & 5601 & 39.3 & 5265 & 82.5 & 3828 & 87.8 & 2696 & 34.8 & 3798 & 57.1 & 4238 \\
      \rowcolor{HighlightRow}
      SHAPE  & \valup{45.6}{7.0} & \valdown{5194}{-26.9\%} 
                & \valup{40.5}{4.8} & \valdown{4896}{-30.5\%} 
                & \valup{84.9}{2.8} & \valdown{3549}{-26.0\%} 
                & \valup{89.0}{1.9} & \valdown{2069}{-34.7\%} 
                & \valup{36.9}{2.4} & \valdown{3115}{-37.3\%} 
                & \valup{59.4}{3.8} & \valdown{3765}{-30.5\%} \\
      \midrule
      
      \rowcolor{ModelGray}
      \multicolumn{13}{l}{\textbf{\textit{Qwen3-4B}}} \\
      GRPO            & 71.3 & 13541 & 65.8 & 15279 & 92.7 & 8051 & 94.0 & 5116 & 48.1 & 6264 & 74.4 & 9650 \\
      MRT             & 70.4 & 12455 & 66.3 & 13691 & 93.1 & 6508 & 93.4 & 3948 & 47.8 & 4874 & 74.2 & 8295 \\
      \rowcolor{HighlightRow}
      SHAPE  & \valup{73.9}{2.6} & \valdown{11028}{-18.6\%} 
                & \valup{67.1}{1.3} & \valdown{12733}{-16.7\%} 
                & \valup{96.8}{4.1} & \valdown{5866}{-27.1\%} 
                & \valup{95.6}{1.6} & \valdown{3338}{-34.8\%} 
                & \valup{54.3}{6.2} & \valdown{4054}{-35.3\%} 
                & \valup{77.5}{3.1} & \valdown{7404}{-23.3\%} \\
      
      \bottomrule
      \end{tabular}%
  }
  \caption{Main results on mathematical reasoning benchmarks. We report Pass@1 accuracy (\%) and average generated token counts across five datasets. \textcolor{GreenText}{Blue} indicates improvement (Acc $\uparrow$ or Tokens $\downarrow$) of SHAPE over GRPO. Compared to GRPO and MRT baselines, {SHAPE} consistently establishes a new Pareto frontier across all three base models, achieving superior accuracy while significantly reducing token consumption.}
  \label{tab:main_results}
  \end{table*}

\subsection{Token-Level Credit Redistribution}
\label{sec:token_redistribution}

While segment-level advantages provide a strategic signal, applying a uniform $A_k$ to all tokens ignores the varying information density within a reasoning step. To capture fine-grained contributions, recent works such as \citet{cheng2025reasoning} and GTPO \citep{tan2025gtpo} have introduced entropy-based reward shaping. However, these methods operate at the \textit{trajectory level}, modulating the high-variance global outcome reward based on token entropy relative to the entire sequence.

SHAPE adapts this insight into a hierarchical context, performing credit redistribution strictly within the local segment. This local scope offers the fundamental advantage of \textit{stable anchoring}. Global outcome rewards are inherently sparse and noisy; modulating them token-wise often amplifies variance. In contrast, SHAPE anchors redistribution to the segment advantage $A_k$—a dense, low-variance signal derived from potential estimation. Floating token rewards around this stable local baseline ensures that we refine a valid signal rather than amplifying chaos.

\paragraph{Standardized Importance \& Modulation.}
We quantify the relative importance of token $x_t$ using a Z-score standardization within its segment. Let $\mu_H$ and $\sigma_H$ be the statistics of the valid entropy sequence in segment $s_k$. The importance weight $w_t$ is computed via a centered affine transformation:
\begin{equation}
\begin{split}
    \tilde{H}(x_t) &= \frac{H(x_t) - \mu_H}{\sigma_H + \epsilon}, \\
    w_t &= \text{clip}\big(1 + \beta \cdot \tilde{H}(x_t), \delta_{\min}, \delta_{\max}\big).
\end{split}
\end{equation}
The final token advantage is obtained by modulating the segment anchor: $A_t = A_k \cdot w_t$. This ensures that tokens with average entropy retain the original segment advantage ($w_t \approx 1$), while pivotal high-entropy decisions receive amplified credit proportional to their local significance.

  \section{Experiments}
  \subsection{Experimental Setup}
  
  Experiments are conducted across three backbone models: DeepSeek-R1-Distill-Qwen-1.5B, DeepScaleR-1.5B-Preview, and Qwen3-4B, using rStar2A \cite{shang2025rstar} as the training dataset. To ensure stability, we employ the \textit{clip-higher} mechanism from DAPO \cite{yu2025dapo}, setting $\epsilon_{\text{high}} = 0.28$ and $\epsilon_{\text{low}} = 0.2$. Unless otherwise noted, the process reward coefficient is uniformly set to $\alpha=0.3$, and the discount factor lower bound is set to $\gamma_{\text{min}}=0.9$. Evaluation spans five standard benchmarks (AIME 2024/25, AMC 2023, MinervaMATH, MATH500), reporting average accuracy and token usage based on consistent sampling parameters ($T=0.6, p=0.95, k=40$, max\_len=32,768). Further details are provided in Appendix \ref{app:main_result}.

  \begin{figure}[t]
    \centering
    \begin{subfigure}[b]{0.49\linewidth}
      \centering
      \includegraphics[width=\linewidth]{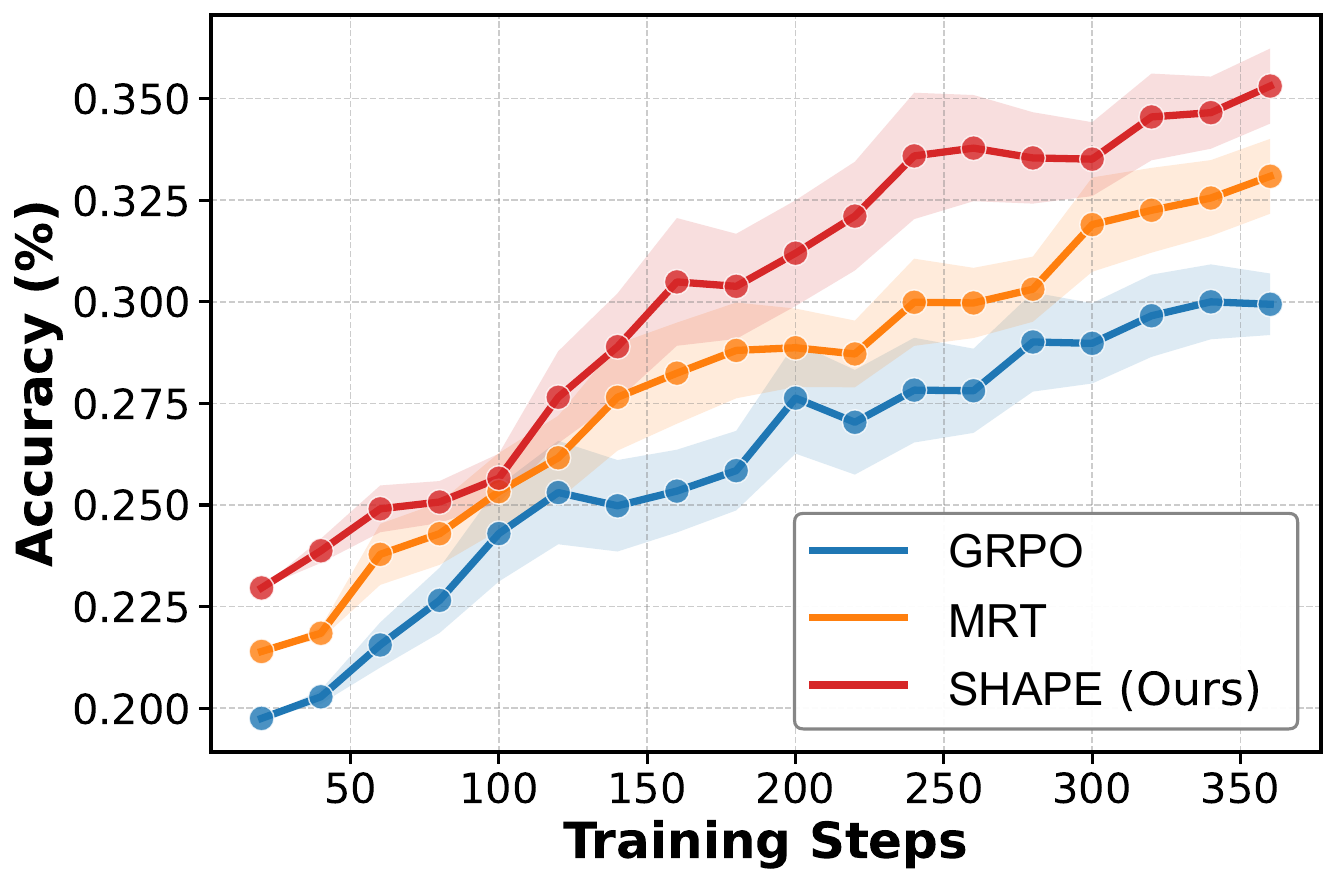} 
      \caption{AIME 24 Accuracy}
      \label{fig:curve_acc}
    \end{subfigure}
    \hfill 
    \begin{subfigure}[b]{0.49\linewidth}
      \centering
      \includegraphics[width=\linewidth]{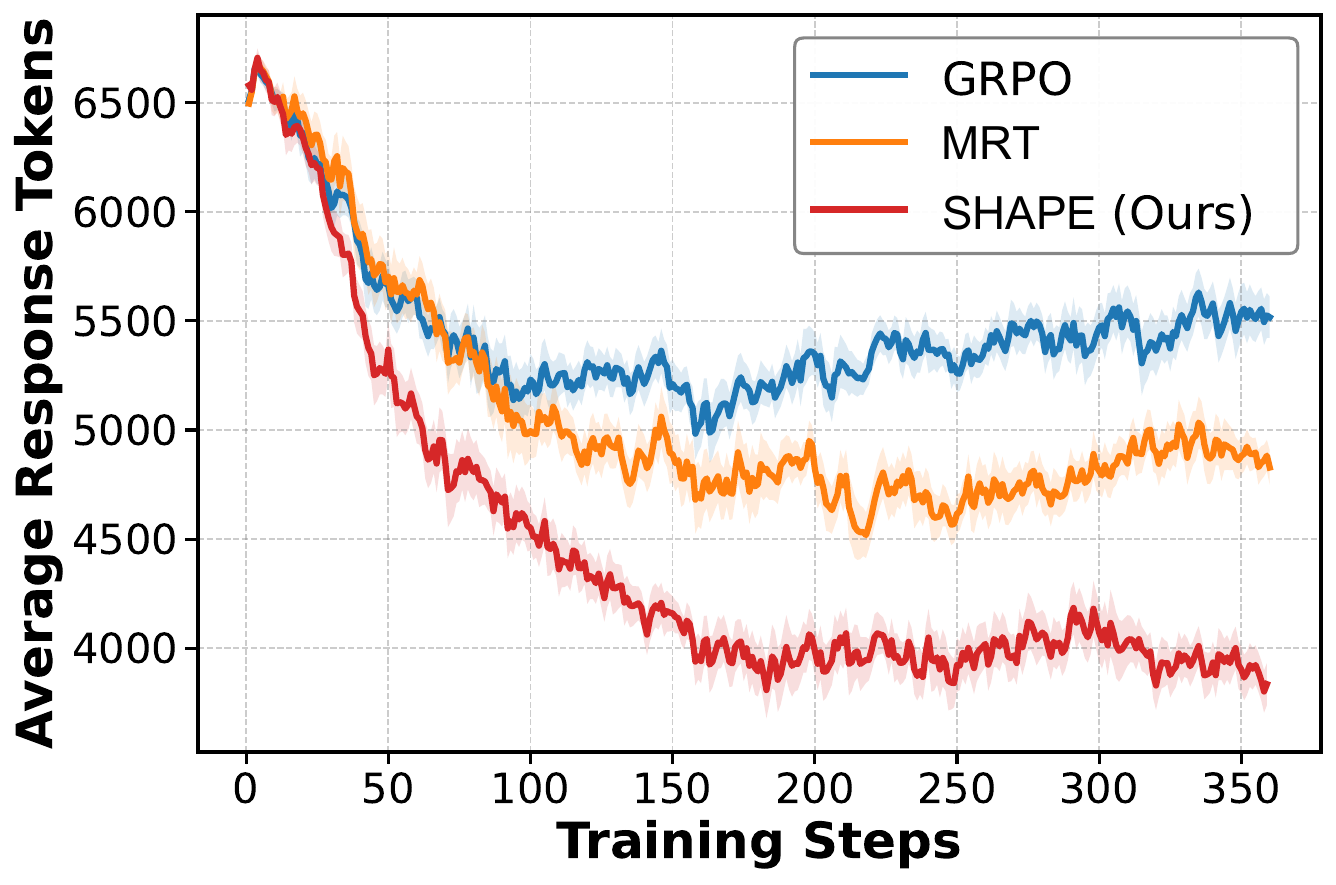}
      \caption{Avg Response Tokens}
      \label{fig:curve_tokens}
    \end{subfigure}
    \caption{Performance of DS-R1-Distill-Qwen-1.5B.}
    \label{fig:training_curves}
  \end{figure}

  \subsection{Main Results}
  \label{sec:main_results}
  
  Table \ref{tab:main_results} demonstrates that SHAPE consistently outperforms both outcome-supervised (GRPO) and static process-supervised (MRT) baselines. SHAPE {simultaneously maximizes reasoning accuracy and token efficiency} across all benchmarks, achieving a substantial reduction in token usage (averaging $\sim${30\%}). 
  Crucially, unlike static approaches that often struggle to balance these two objectives, SHAPE's stage-aware shaping ($\gamma_k$) successfully compresses reasoning paths while preserving—and often enhancing—semantic integrity. 
  This dual improvement is further corroborated by Figure \ref{fig:training_curves}, where SHAPE maintains a clear lead in test accuracy while driving a steep reduction in response length throughout training. To further verify that these gains reflect enhanced general reasoning rather than narrow domain overfitting, we evaluate on out-of-distribution benchmarks in Appendix~\ref{app:ood}.

\subsection{Ablation Study}
\label{sec:ablation}

Table \ref{tab:ablation} summarizes the contribution of SHAPE's components and parameter sensitivity on DeepSeek-R1-Distill-Qwen-1.5B model.

\paragraph{Impact of Core Components.}
We isolate the effects of Entropy-Based Segmentation (EBS) and Token-Level Credit Redistribution (TCR).
\begin{itemize}
    \item \textbf{w/o EBS:} The performance decline validates that EBS effectively delineates semantic units. Unlike rigid heuristics (e.g., "\texttt{\textbackslash n\textbackslash n}"), EBS aligns segmentation with logical boundaries, minimizing noise in potential estimation.
    \item \textbf{w/o TCR:} Removing TCR causes a notable accuracy drop, confirming its necessity. By amplifying high-entropy tokens, TCR incentivizes effort at pivotal steps, preventing the model from defaulting to shallow, safe paths.
\end{itemize}

\begin{table}[t]
  \centering

  \small 
  \setlength{\tabcolsep}{2pt} 
  
  \begin{tabular}{lllll}
  \toprule
  \multirow{2}{*}{\textbf{Method}} & 
  \multicolumn{2}{c}{\textbf{AIME 24}} & 
  \multicolumn{2}{c}{\textbf{AIME 25}} \\
  \cmidrule(lr){2-3} \cmidrule(lr){4-5}
   & Acc & Tokens & Acc & Tokens \\
  \midrule
  
  GRPO & 34.7 & 8772 & 27.5 & 8109 \\
    
  \rowcolor{HighlightRow}
  \textbf{SHAPE (Ours)} \quad & \textbf{37.1} & \textbf{6164} & \textbf{31.8} & \textbf{5425} \\
  \midrule

  \multicolumn{5}{l}{\textit{Component Analysis}} \\

  w/o EBS & \accdown{36.8}{0.3} & \tokup{6380}{3.5\%} & \accdown{31.6}{0.2} & \tokup{5590}{3.0\%} \\

  w/o TCR & \accdown{36.2}{0.9} & \tokdown{6080}{1.4\%} & \accdown{29.8}{2.0} & \tokdown{5250}{3.2\%} \\
  \midrule

  \multicolumn{5}{l}{\textit{Parameter Sensitivity ($\gamma$)}} \\

  Fixed $\gamma_k=0.9$      & \accdown{36.5}{0.6} & \tokup{6955}{12.8\%} & \accup{32.5}{0.7} & \tokup{6610}{21.8\%} \\

  $\gamma_{\min}=0.95$      & \accup{37.6}{0.5} & \tokup{6340}{2.9\%} & \accdown{30.7}{1.1} & \tokup{5769}{6.3\%} \\

  $\gamma_{\min}=0.8$       & \accdown{36.3}{0.8} & \tokdown{5720}{7.2\%} & \accdown{30.9}{0.9} & \tokdown{5010}{7.6\%} \\
 
  $\gamma_{\min}=0.7$       & \accdown{30.8}{6.3} & \tokdown{4580}{25.7\%} & \accdown{26.2}{5.6} & \tokdown{3920}{27.7\%} \\

  \bottomrule
  \end{tabular}
    \caption{Ablation study. Subscripts indicate gaps relative to SHAPE (\textcolor{GreenText}{blue}: improvement; \textcolor{RedText}{red}: degradation).}
  \label{tab:ablation}
\end{table}

\paragraph{Sensitivity to Dynamic Discounting ($\gamma$).}
We analyze the length-dependent discount factor $\gamma_k$.
\begin{itemize}
    \item \textbf{Fixed $\gamma_k=0.9$:} The surge in token usage confirms that static discounting fails to impose progressive efficiency constraints, validating the necessity of our length-dependent design in curbing verbosity.
    \item \textbf{Varying $\gamma_{\min}$:} A balanced decay is crucial. Relaxing the constraint ($\gamma_{\min}=0.95$) inflates computational cost, while over-aggressive discounting ($\gamma_{\min}=0.7$) leads to performance collapse. The latter indicates that excessive penalties force the model to prematurely truncate reasoning to avoid the length tax, rather than solving the problem. A formal theoretical derivation of the critical lower bound for $\gamma_{\min}$ is provided in Appendix~\ref{app:gamma_bound}.
\end{itemize}

\section{Analysis}

Unless otherwise specified, all analytical experiments in this section are conducted using the {DeepSeek-R1-Distill-Qwen-1.5B} model.

\subsection{Sensitivity Analysis of Potential Gains}
\label{sec:sensitivity_analysis}

To validate the {Stage Awareness} principle proposed in the Introduction—specifically that breakthroughs from low-potential states are more valuable—we analyze the correlation between immediate potential gain ($\Delta$) and final success, stratified by starting potential $\Phi(s_k)$ (see Appendix \ref{app:sensitivity_details} for statistical details).

\begin{figure}[t]
  \centering
  \includegraphics[width=\linewidth]{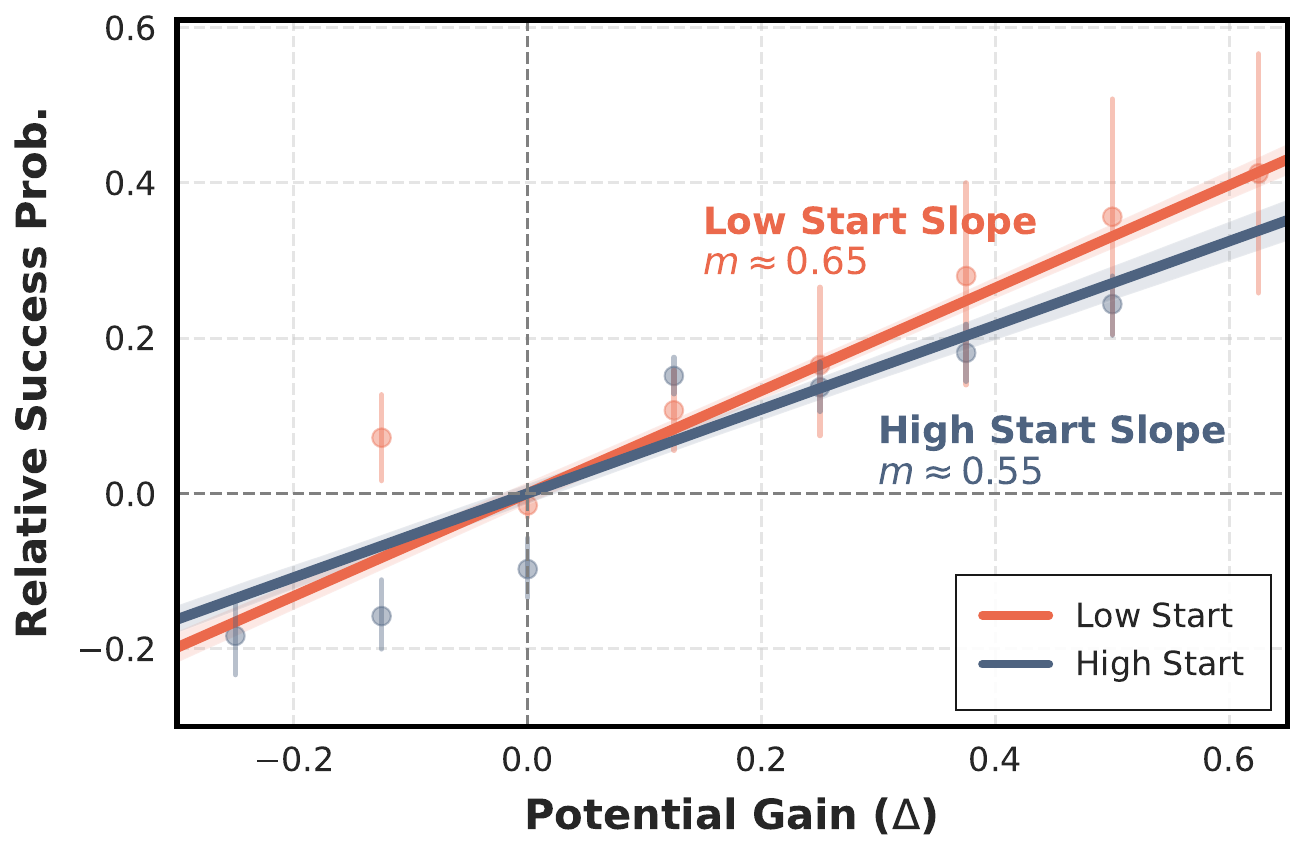} 
  \caption{{Marginal effect of potential gains.} The steeper regression slope for the Low Start group confirms that improvements in adverse states are more critical for final success.}
  \label{fig:sensitivity}
\end{figure}

As shown in Figure \ref{fig:sensitivity}, the {Low Start} group ($\Phi \le 0.25$) exhibits a significantly steeper regression slope ($k \approx 0.65$) compared to the {High Start} group ($k \approx 0.55$). This empirical gap implies that a unit of improvement in adverse stages yields an approximately {18\% higher marginal return} on final success than in already performant stages. This confirms that rescuing a failing path is far more decisive than refining a successful one, providing strong empirical justification for SHAPE's stage-aware weighting mechanism.

\subsection{Evolution of Reasoning Strategy}
\label{sec:reasoning_strategy}

\begin{figure}[t]
  \centering
  \includegraphics[width=\linewidth]{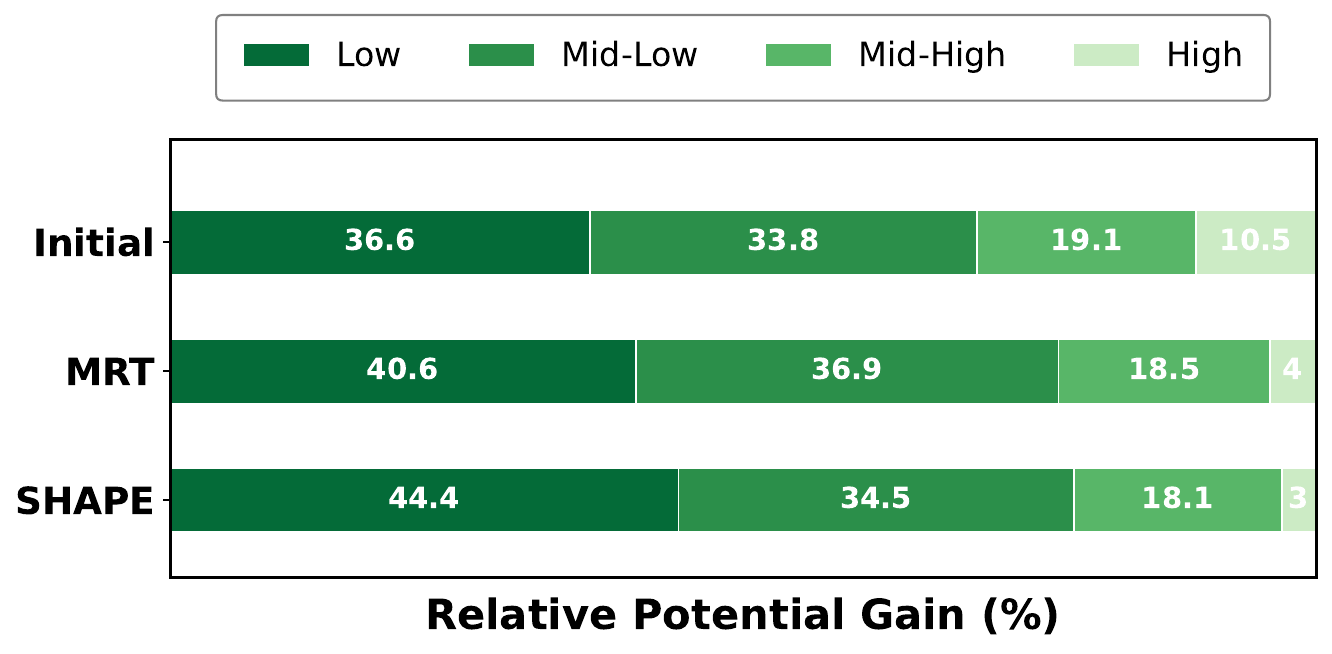} 
\caption{{Distribution of potential gain contributions.} SHAPE shifts the focus towards the \textit{Low Start} regime ($44.4\%$ vs. MRT's $40.6\%$), validating its capability to rectify reasoning paths from poor initial states.}  \label{fig:value_gain_distribution}
\end{figure}

\begin{figure}[t]
  \centering
  \includegraphics[width=\linewidth]{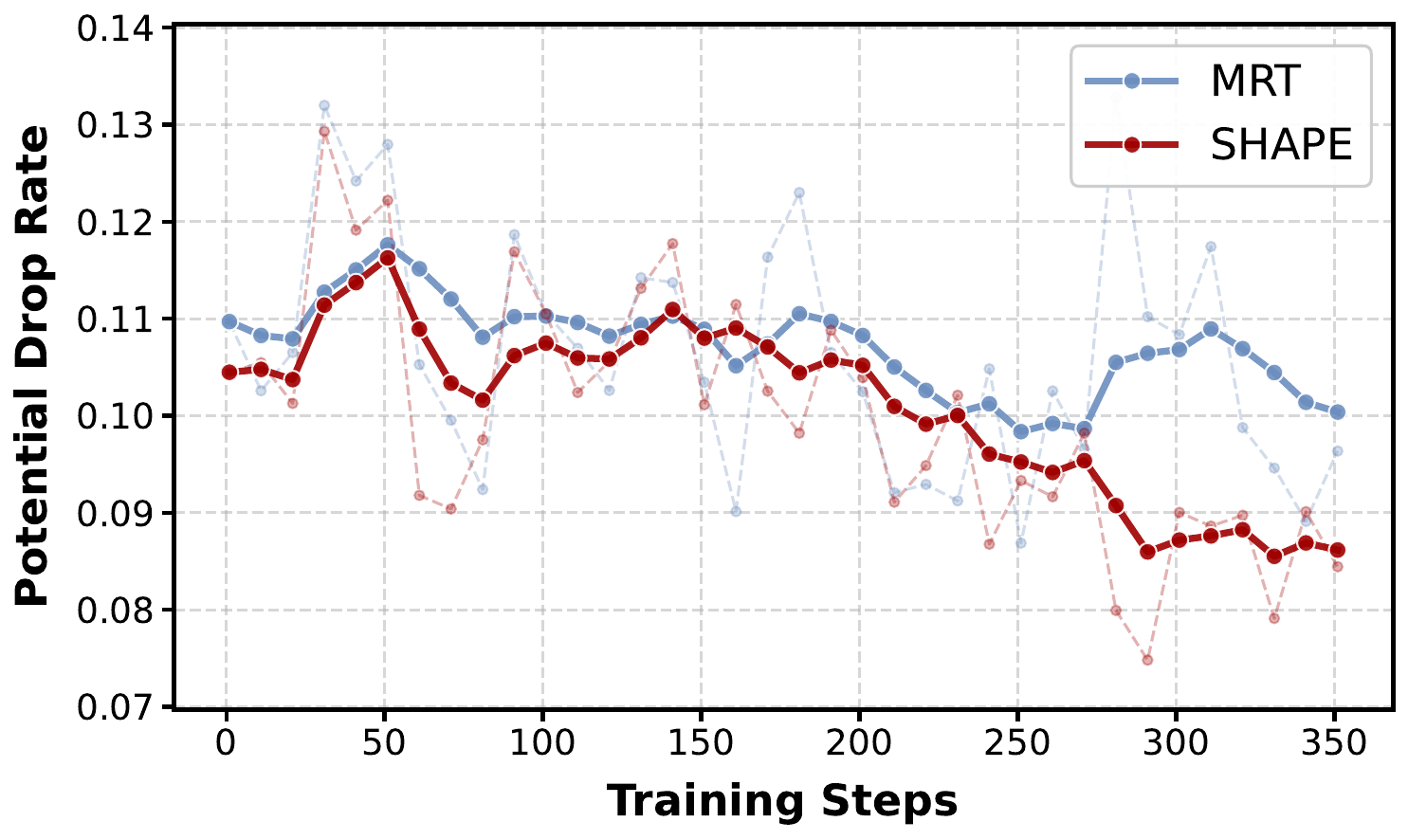} 
  \caption{{Potential drop rate during training.} The curves depict the proportion of adjacent segment transitions where potential decreases ($\Phi(s_{k+1}) < \Phi(s_k)$). }
  \label{fig:value_drop}
\end{figure}

Having established in \S\ref{sec:sensitivity_analysis} that improvements in low-potential stages are more decisive, we now verify if SHAPE effectively aligns its optimization focus with this insight. We analyze the \textit{sources} of realized potential gains—categorized by starting potential—to see where progress stems from (methodology in Appendix \ref{app:value_gain_analysis}).

As visualized in Figure \ref{fig:value_gain_distribution}, SHAPE significantly alters the optimization landscape. It derives the highest proportion of gains from Low Start states ({44.4\%}, compared to MRT's 40.6\%), indicating a behavioral shift toward rectifying adverse situations. Conversely, contributions from High Start states drop to just 3\% (vs. Initial's 10.5\%). This confirms that SHAPE suppresses easy rewards derived from merely maintaining high potentials, forcing the model to tackle the critical, high-difficulty early phases of reasoning.

\subsection{Verification of Strategic Sandbagging}
\label{sandbagging}
To empirically validate the \textit{Sandbagging Risk} hypothesized in \S\ref{subsec:mrt}, we monitor the frequency of negative potential changes ($\Phi(s_{k+1}) < \Phi(s_k)$). As shown in Figure \ref{fig:value_drop}, while initial trends are similar, a distinct divergence emerges. MRT exhibits high volatility with conspicuous late-stage spikes, corroborating that its endpoint-based advantage ($R - \Phi(s_k)$) incentivizes gaming—deliberately traversing low-potential states to inflate subsequent recovery rewards. Conversely, SHAPE maintains a steady, low-variance decline. This confirms that our PBRS-based formulation ($\gamma_k \Phi(s_{k+1}) - \Phi(s_k)$) effectively closes this loophole, strictly penalizes regressive steps and enforces monotonic progress.

\subsection{Impact of Segmentation Granularity}
\label{sec:granularity_impact}

\begin{figure}[t]
  \centering
  \includegraphics[width=0.95\linewidth]{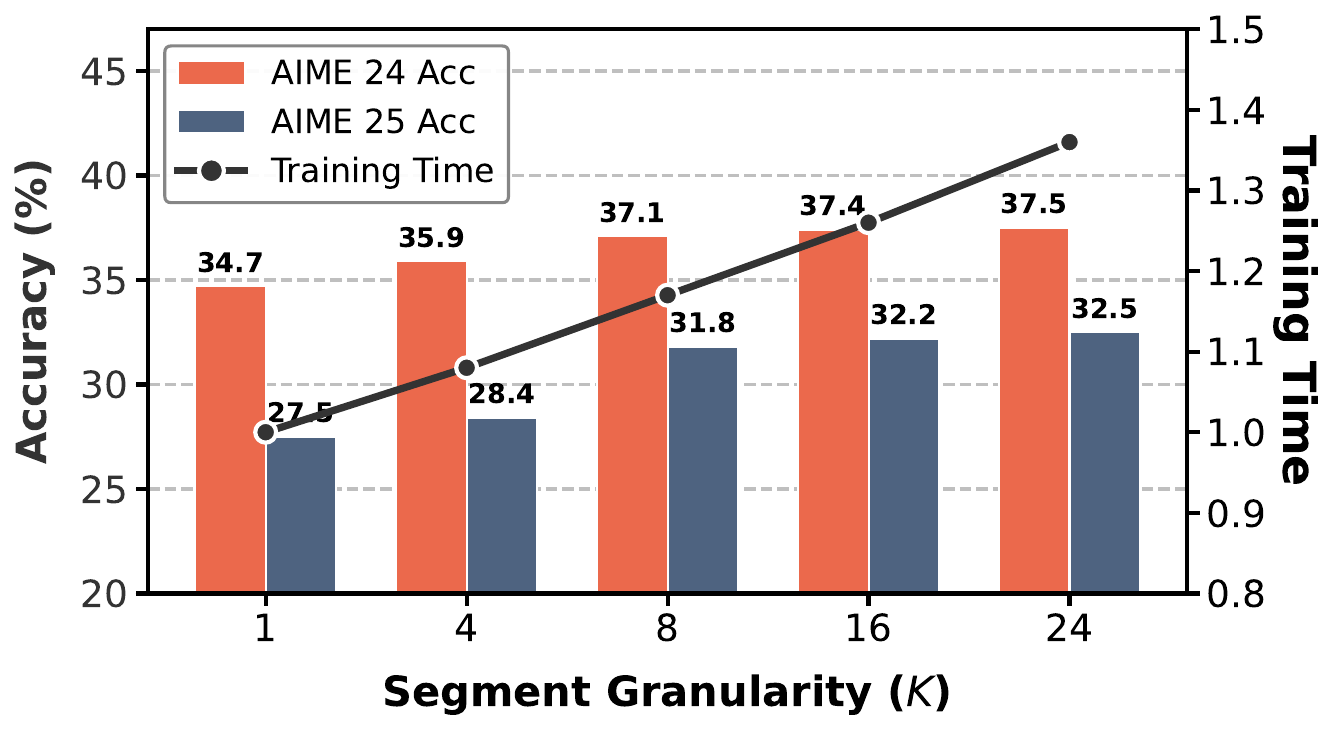} 
\caption{{Granularity trade-off.} Accuracy (lines) saturates after $K=8$, while training cost (bars) scales linearly. $K=8$ represents the optimal balance between performance gains and computational overhead.}  \label{fig:k_ablation}
\end{figure}

We analyze the sensitivity to segment count $K$. Figure \ref{fig:k_ablation} reveals the trade-off between model performance and computational cost.

\paragraph{Diminishing Returns \& The Sweet Spot.}
Increasing $K$ from 1 to 8 yields substantial accuracy gains, confirming that intermediate checkpoints significantly aid potential estimation. However, further increasing $K$ to 16 or 24 results in performance saturation or marginal fluctuations. This indicates a marginal utility threshold: $K=8$ provides sufficient resolution to capture reasoning pivots, whereas excessive segmentation ($K > 8$) risks over-fragmenting semantic units without adding meaningful signal.

\paragraph{Training Cost vs. Inference Saving.}

While training cost scales linearly with $K$, the investment is strategically justified. Specifically, $K=8$ incurs a manageable $1.17\times$ training overhead compared to the baseline ($K=1$), but unlocks the significant ($\sim 30\%$) recurring inference savings demonstrated in \S\ref{sec:main_results}. This represents an optimal {Total Cost of Ownership (TCO)} trade-off: swapping a marginal increase in one-time training compute for substantial, permanent deployment efficiency.

\begin{figure}[t]
  \centering
  \includegraphics[width=\linewidth]{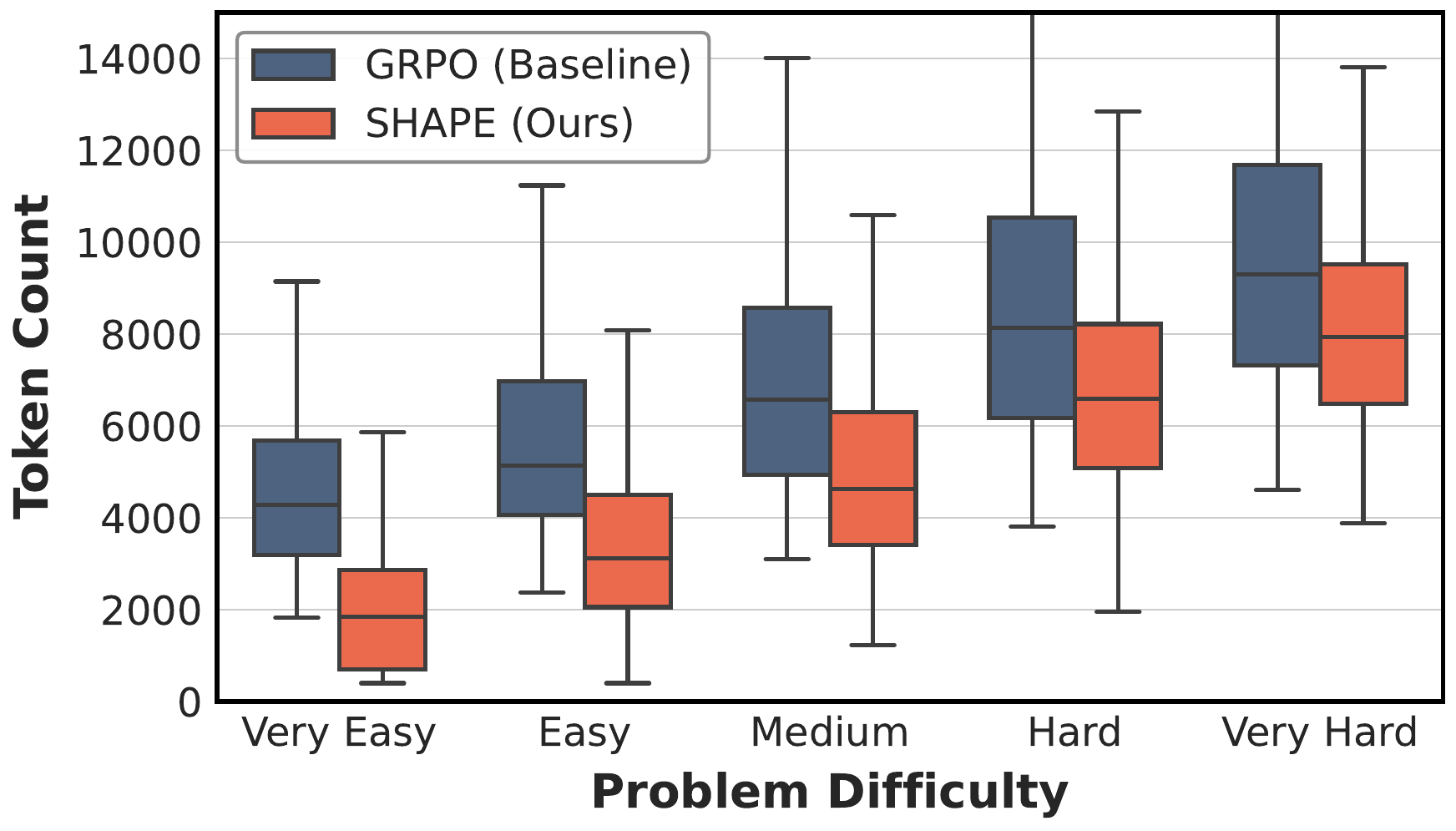} 
\caption{{Response length vs. Problem difficulty.} SHAPE exhibits a steeper scaling slope and lower variance compared to GRPO, indicating precise resource allocation based on problem hardness.}  \label{fig:adaptive_scaling}
\end{figure}

\subsection{Analysis of Adaptive Computation}
\label{sec:adaptive_computation}

We investigate SHAPE's ability to dynamically align reasoning cost with problem complexity. We categorize test problems into five difficulty bins and analyze the distribution of response lengths (methodology in Appendix~\ref{app:adaptive_scaling}).

As shown in Figure \ref{fig:adaptive_scaling}, SHAPE exhibits a steeper scaling slope accompanied by lower variance compared to GRPO. This indicates a robust perception of difficulty: rather than engaging in aimless exploration, SHAPE precisely targets the reasoning depth required for a specific hardness level. Furthermore, SHAPE consistently maintains lower token counts across all bins. This validates that the dynamic discount factor $\gamma_k$ effectively eliminates redundancy, ensuring the model extends its reasoning chain only when necessary.

\subsection{Analysis of Token Length Distribution}
\label{app:length_distribution}

In this section, we analyze the distribution of token usage for generating solutions across three representative benchmarks: AIME 2025 (Hard), MinervaMATH (Medium), and MATH500 (Easy).

\begin{figure*}[ht]
  \centering
  \begin{subfigure}[b]{0.32\textwidth}
      \centering
      \includegraphics[width=\linewidth]{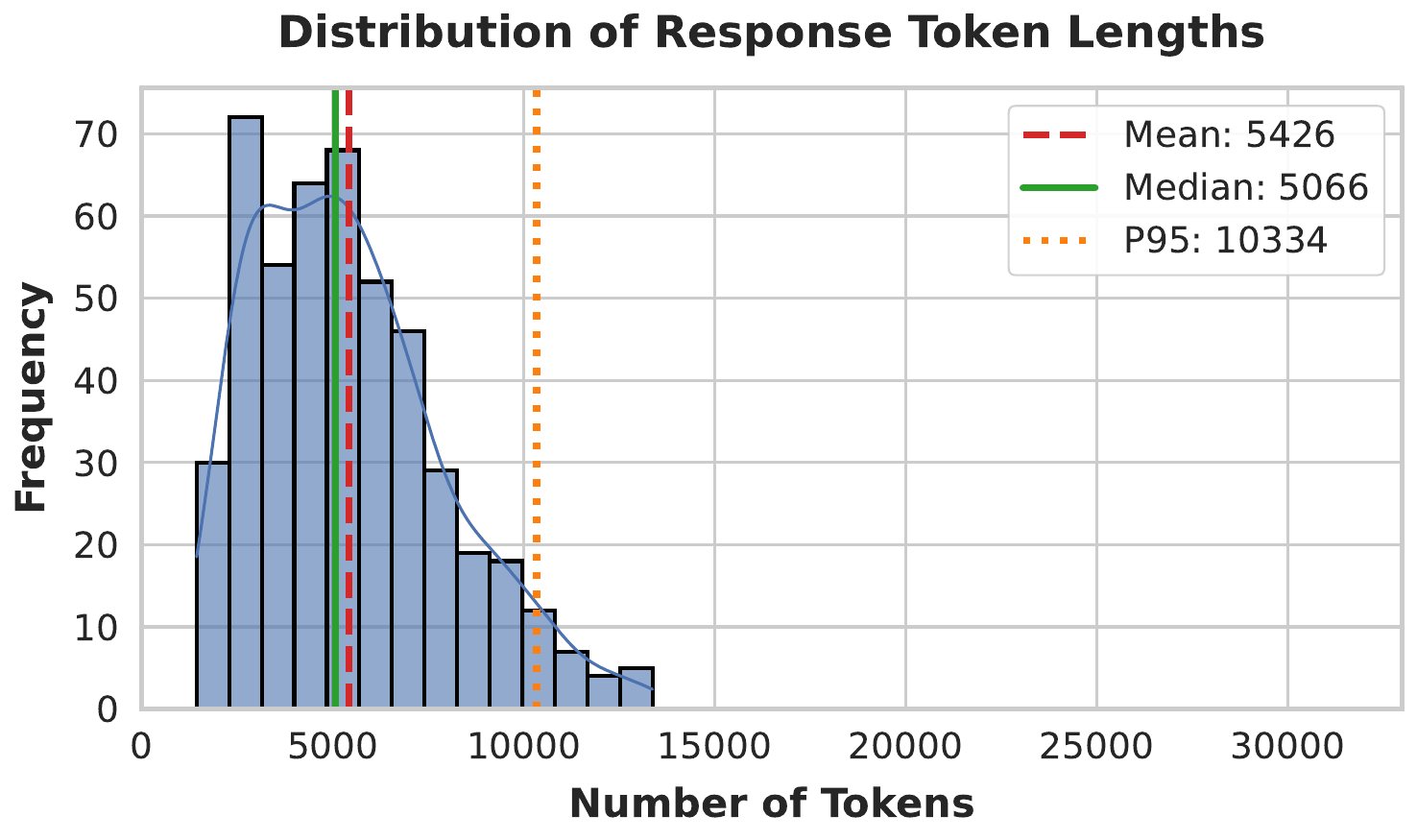} 
      \caption{SHAPE on AIME 2025}
      \label{fig:shape_aime}
  \end{subfigure}
  \hfill
  \begin{subfigure}[b]{0.32\textwidth}
      \centering
      \includegraphics[width=\linewidth]{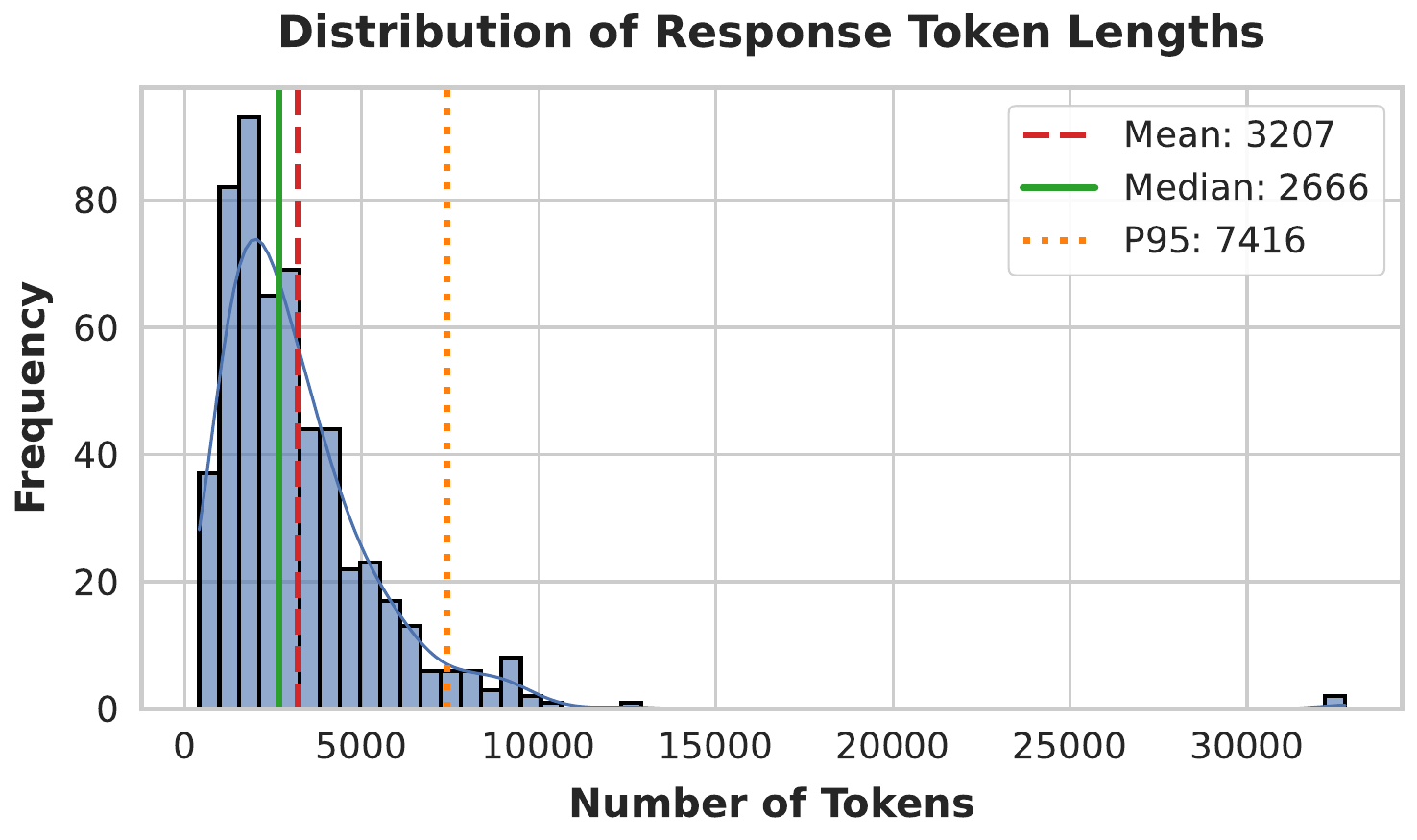}
      \caption{SHAPE on MinervaMATH}
      \label{fig:shape_minerva}
  \end{subfigure}
  \hfill
  \begin{subfigure}[b]{0.32\textwidth}
      \centering
      \includegraphics[width=\linewidth]{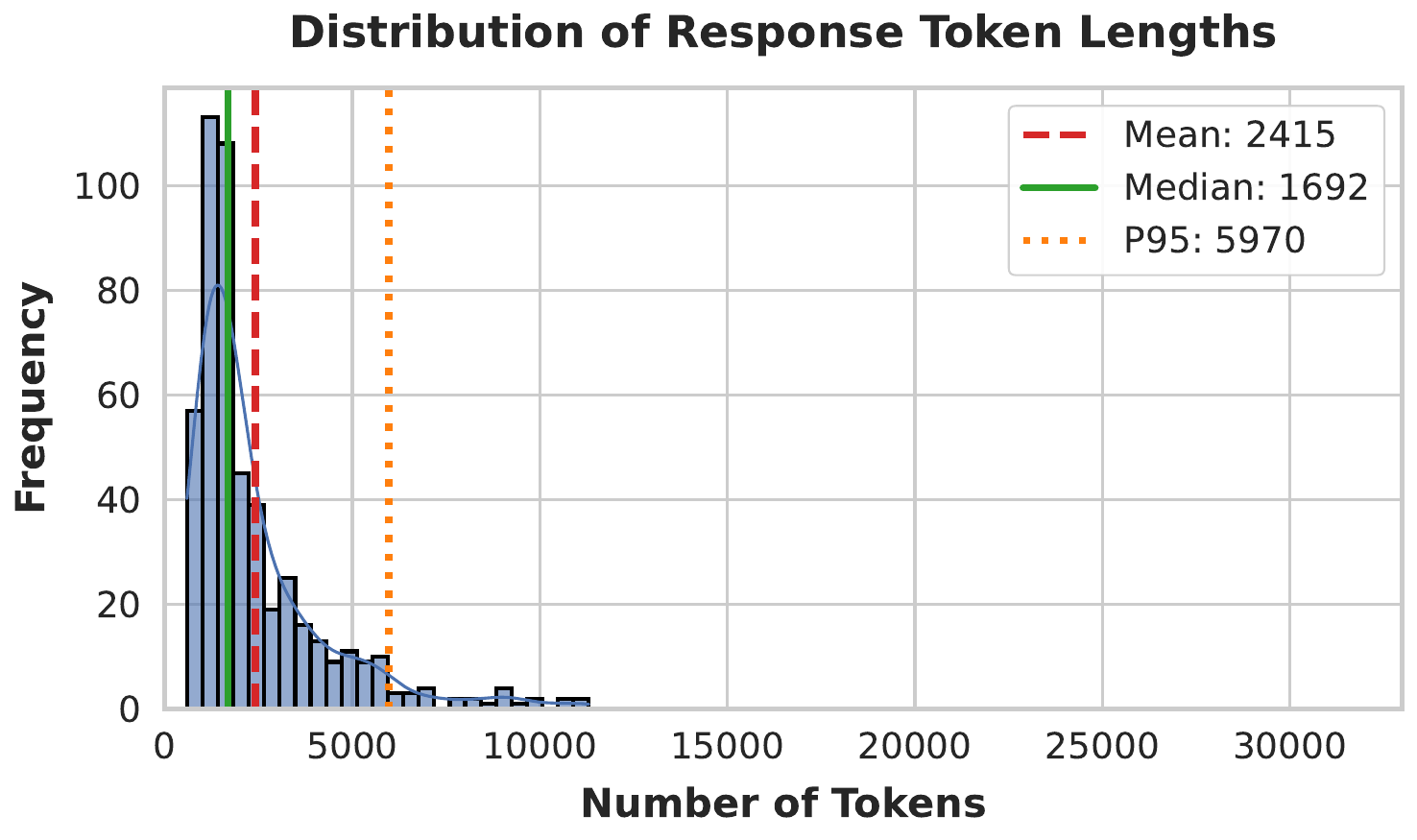}
      \caption{SHAPE on MATH500}
      \label{fig:shape_math500}
  \end{subfigure}
  
  \vspace{0.3cm} 

  \begin{subfigure}[b]{0.32\textwidth}
    \centering
    \includegraphics[width=\linewidth]{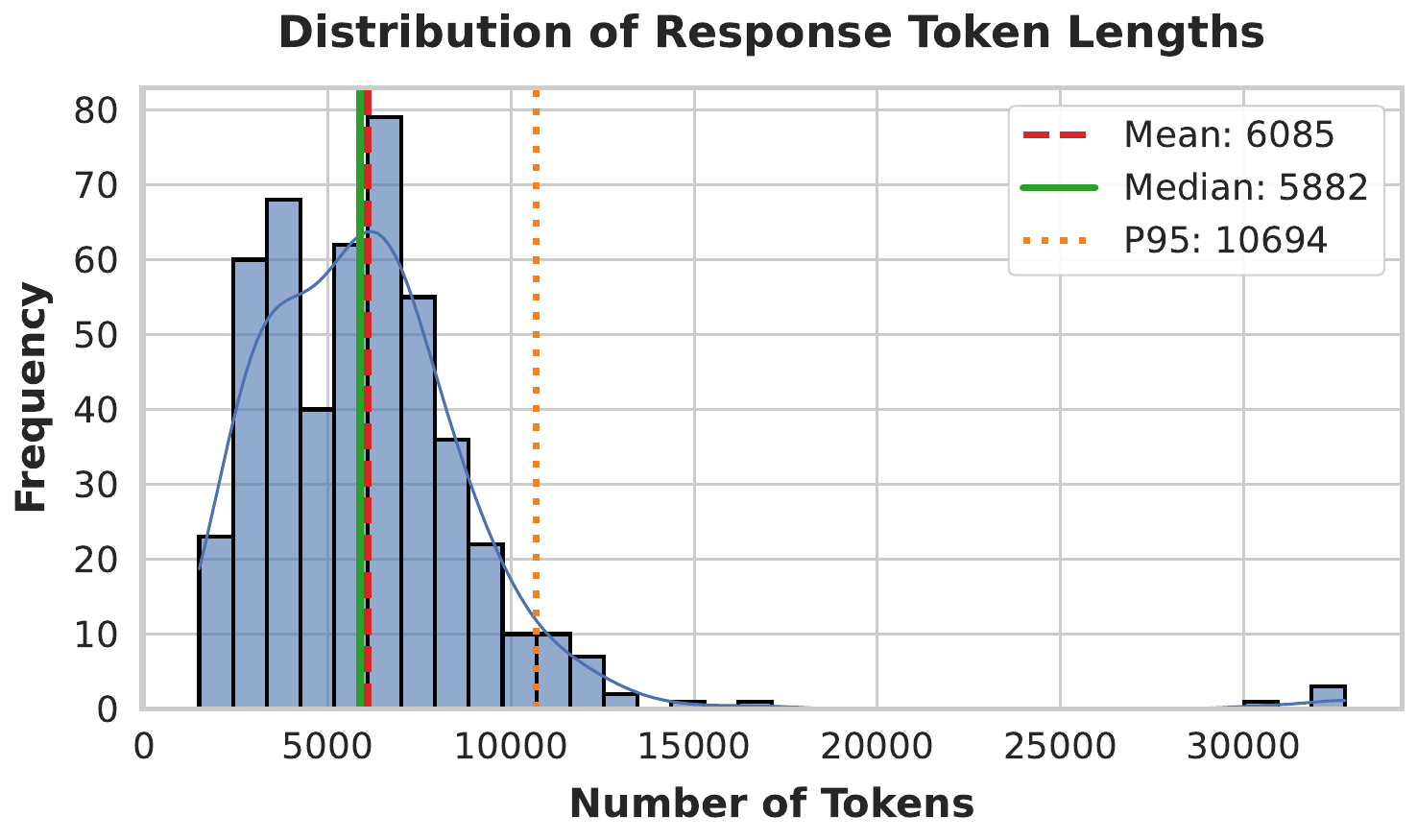}
    \caption{MRT on AIME 2025}
    \label{fig:mrt_aime}
\end{subfigure}
\hfill
\begin{subfigure}[b]{0.32\textwidth}
    \centering
    \includegraphics[width=\linewidth]{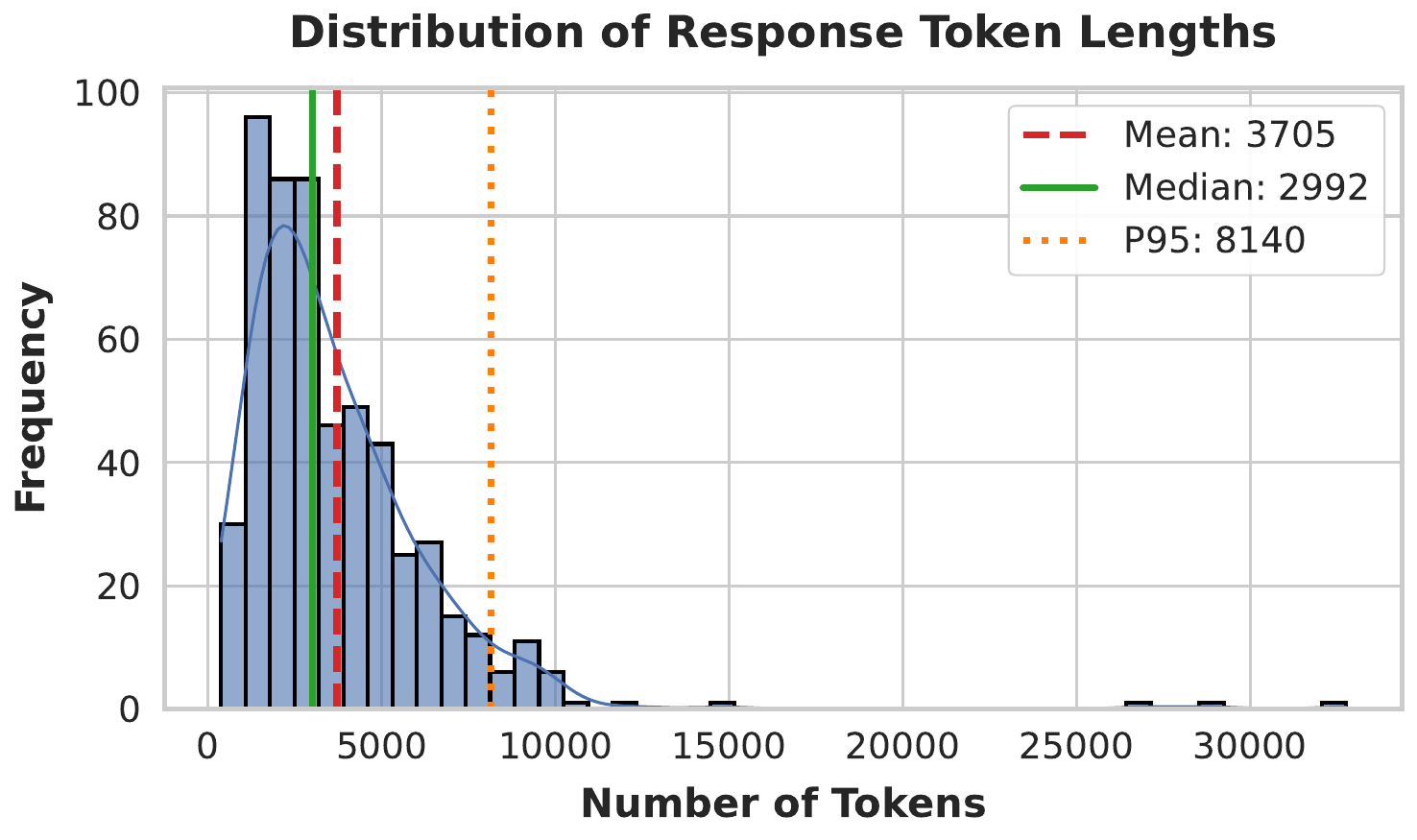}
    \caption{MRT on MinervaMATH}
    \label{fig:mrt_minerva}
\end{subfigure}
\hfill
\begin{subfigure}[b]{0.32\textwidth}
    \centering
    \includegraphics[width=\linewidth]{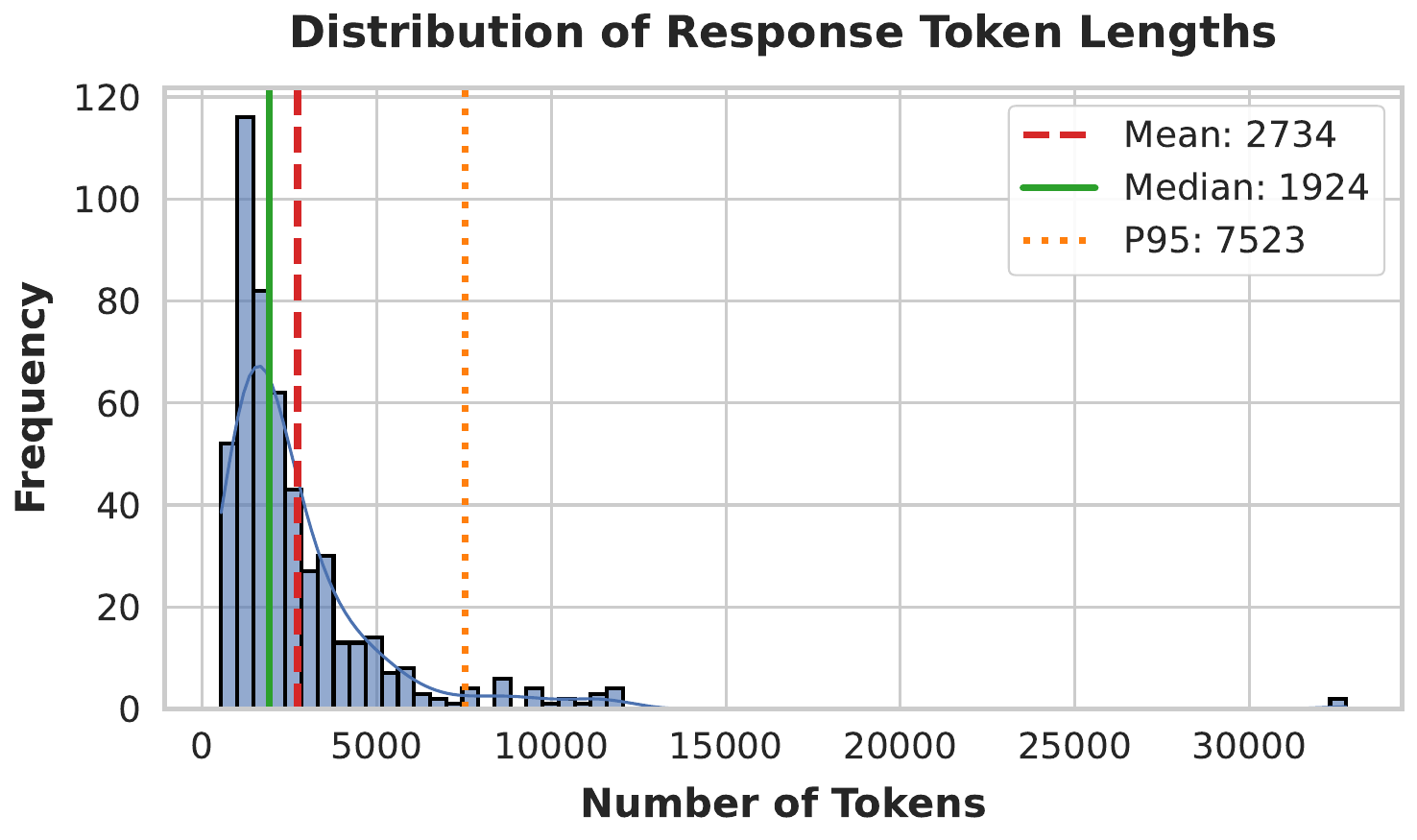}
    \caption{MRT on MATH500}
    \label{fig:mrt_math500}
\end{subfigure}

  \begin{subfigure}[b]{0.32\textwidth}
      \centering
      \includegraphics[width=\linewidth]{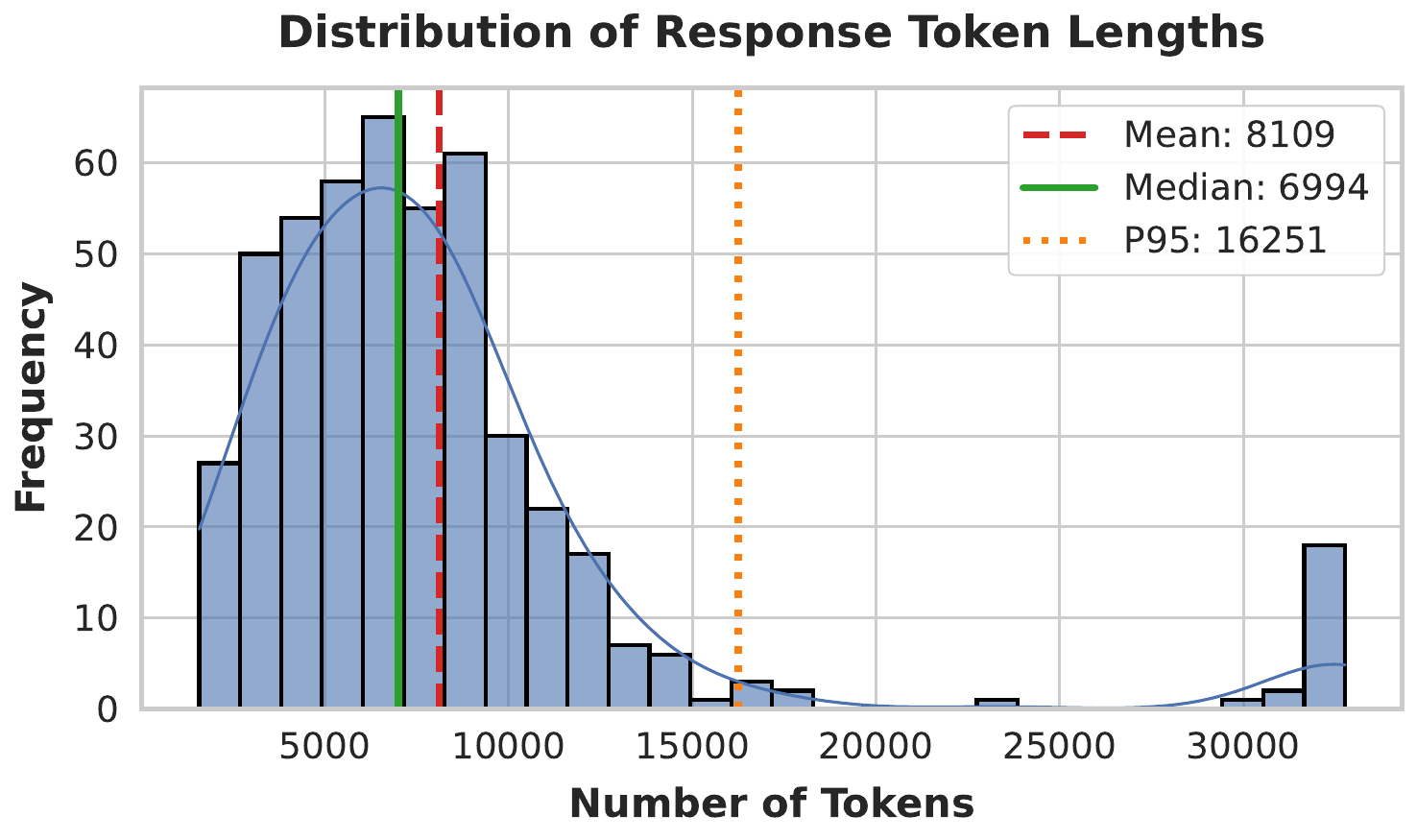}
      \caption{GRPO on AIME 2025}
      \label{fig:grpo_aime}
  \end{subfigure}
  \hfill
  \begin{subfigure}[b]{0.32\textwidth}
      \centering
      \includegraphics[width=\linewidth]{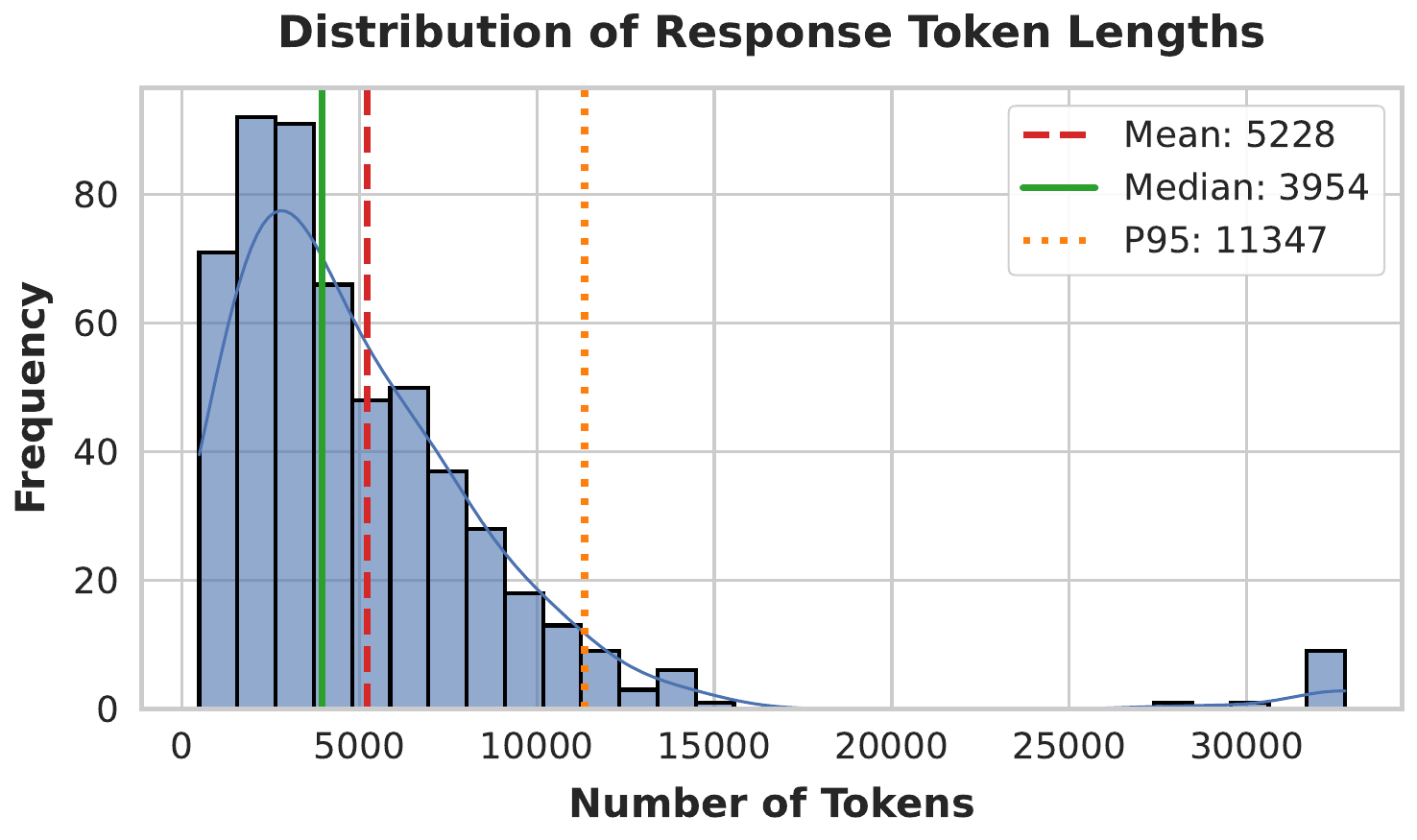}
      \caption{GRPO on MinervaMATH}
      \label{fig:grpo_minerva}
  \end{subfigure}
  \hfill
  \begin{subfigure}[b]{0.32\textwidth}
      \centering
      \includegraphics[width=\linewidth]{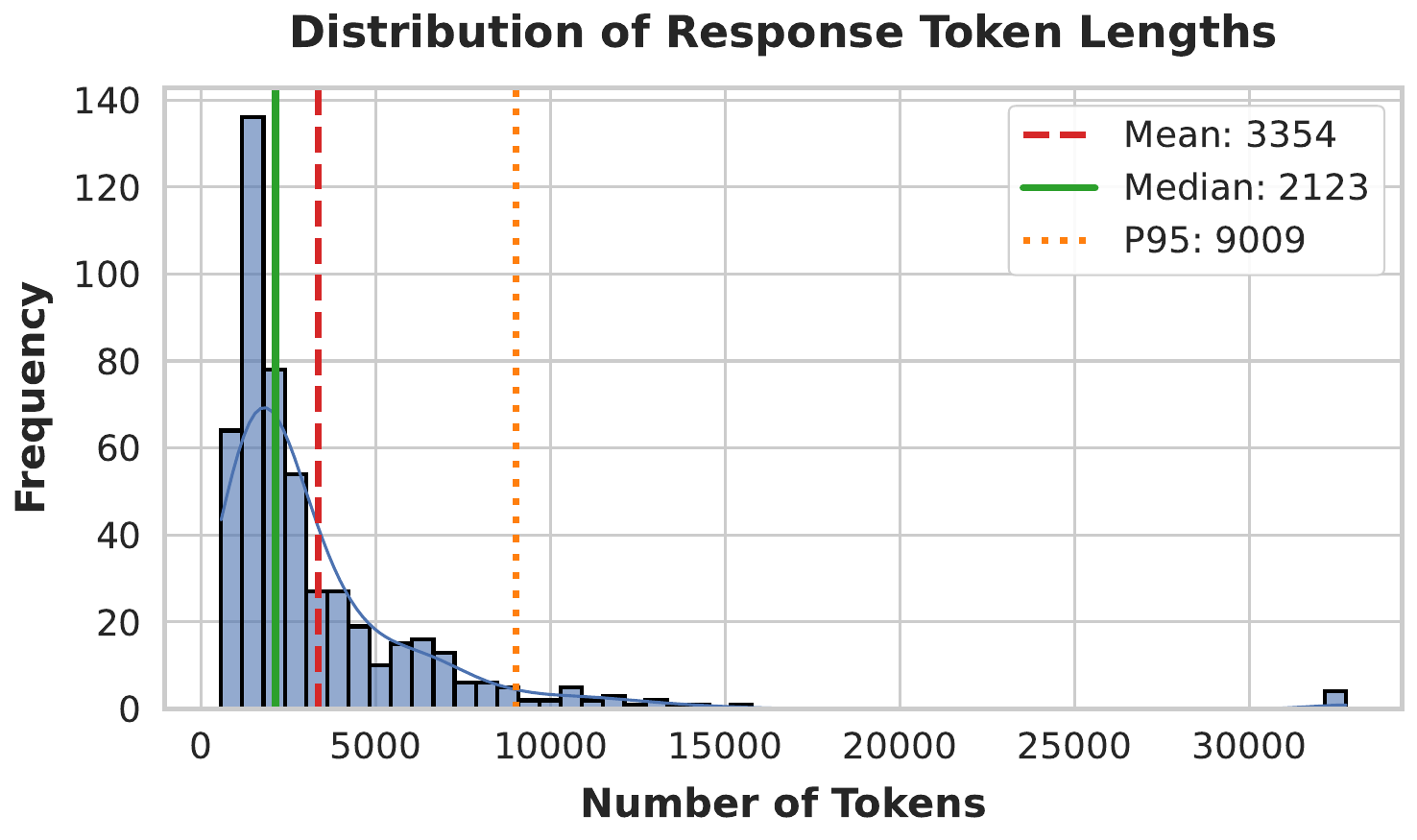}
      \caption{GRPO on MATH500}
      \label{fig:grpo_math500}
  \end{subfigure}
  
  \caption{{Token length distributions.} SHAPE (top) shows a smooth long-tail distribution, while GRPO (bottom) exhibits anomalous spikes near the 32k context limit, indicating degenerate behavior on hard problems. MRT (middle) reduces but does not eliminate such spikes.}
  \label{fig:length_distribution_grid}
\end{figure*}

The distribution plots (Figure \ref{fig:length_distribution_grid}) were generated using Kernel Density Estimation (KDE) overlaying standard histograms, with curves representing the estimated probability density of response lengths.

\paragraph{1. Skewed Long-Tail Distributions.}
All three models exhibit a right-skewed, long-tail distribution: the peak density is concentrated at shorter lengths, indicating that most problems are solved within a reasonable token budget, while the tail reflects longer chains on more complex queries.

\paragraph{2. Mitigation of Reasoning Collapse.} 
A critical anomaly is observed in the GRPO baseline, particularly on harder benchmarks like AIME 2025 and MinervaMATH. There is a distinct {"spike"} or sudden surge in frequency near the maximum generation limit (32k tokens). This phenomenon typically signifies {reasoning collapse}, where the model, unable to find a solution path for extremely hard problems, enters a degenerate state of repetitive loops or incoherent babbling until it hits the hard cutoff.

MRT shows a marked reduction in these spikes compared to GRPO, as its dense feedback partially alleviates blind search on hard problems; however, degenerate responses still persist, indicating that static supervision lacks the mechanism to efficiently terminate low-value paths. In contrast, SHAPE largely eliminates such spikes across all difficulty levels, with curves decaying smoothly to zero well before the limit. This validates that the length-aware discount factor ($\gamma_k$) creates an effective reasoning tax, forcing early termination on dead-end paths and preventing futile context stuffing.

\section{Conclusion}

In this work, we presented {SHAPE} to address the critical limitations of existing process supervision. Grounded in the landscape of empirical solvability, SHAPE implements a \textit{hierarchical credit assignment} strategy that harmonizes reasoning capability with efficiency. At the segment level, it employs stage-aware potential shaping to distinguish meaningful breakthroughs from mere verbosity; at the token level, it utilizes entropy-driven redistribution to sharpen execution signals. Extensive experiments across five benchmarks confirm that SHAPE establishes a superior {Pareto frontier}: achieving an average accuracy gain of {3\%} with {30\%} reduced token consumption. These results validate SHAPE as a robust paradigm for efficient LLM reasoning.

\clearpage

\section*{Limitations}

Our current investigation focuses primarily on {mathematical reasoning tasks} characterized by deterministic verifiability. This deliberate scoping allows for precise calibration of the potential function $\Phi$, as the binary nature of correctness provides a rigorous anchor for estimating solvability without ambiguity. While the core principles of SHAPE are theoretically transferable to broader contexts, extending them to open-ended domains with subjective evaluation criteria (e.g., creative writing or code generation) presents distinct challenges in quantifying progress, thus remaining an exciting avenue for future exploration.

\section*{Ethics Statement}

This work studies LLM reasoning on mathematical tasks using publicly available models and benchmarks for research purposes only. We identify no additional ethical risks beyond those documented by the original dataset creators.

\section*{Acknowledgements}

We thank Yuan Zhou, Hongye Zhou, Yingtong Hu, and Yue Gong for their generous support that made this work possible. We also thank Linyu Liu, Tao Xiao, and Xiaobing Du for their valuable feedback and suggestions on this paper. Finally, we thank the anonymous reviewers for their constructive comments.

\bibliography{acl_latex}

\clearpage

\appendix
\section{Theoretical Analysis}

\subsection{Theoretical Analysis of Task Consistency}
In this section, we provide a mathematical analysis of the Stage-Aware Discounted Progress (SDP) mechanism. Our primary goal is to prove that the introduction of the dynamic discount factor $\gamma_k(L_k)$ strictly preserves the learning objective through \textit{Strong Task Consistency}.
\label{subsec:task_consistency}

We define the total reward for a trajectory $\tau$ of length $K$ as:
\begin{equation}
  \small
    R_{\text{total}}(\tau) = \sum_{k=1}^K \left( R_{\text{outcome}} + \alpha (\gamma_k \Phi(s_{k+1}) - \Phi(s_k)) \right).
\end{equation}
We operate under two standard assumptions: bounded potentials $\Phi(s) \in [0, 1]$ and bounded discount factors $\gamma_k \in [\gamma_{\min}, 1]$.
{Strong Task Consistency} requires that the minimum reward of any correct trajectory strictly dominates the maximum reward of any incorrect trajectory:
\begin{equation}
    \min_{\tau^+ \in \mathcal{T}^+} R_{\text{total}}(\tau^+) > \max_{\tau^- \in \mathcal{T}^-} R_{\text{total}}(\tau^-)
\end{equation}

\paragraph{1. Upper Bound for Incorrect Trajectories ($\tau^-$).}
For an incorrect trajectory, $R_{\text{outcome}}=0$. The reward depends entirely on the accumulated shaping signal. To find the global maximum (the worst-case for consistency), we expand the summation of the shaping term:
\begin{equation}
  \small 
  \begin{aligned}
      R_{\text{total}}(\tau^-) &= \alpha \sum_{k=1}^{K^-} (\gamma_k \Phi(s_{k+1}) - \Phi(s_k)) \\
      &= \alpha \left[ \Phi(s_{K^-}) - \Phi(s_0) + \sum_{k=1}^{K^--1} (\gamma_k - 1)\Phi(s_k) \right]
  \end{aligned}
  \end{equation}
Since $\gamma_k \le 1$ and $\Phi(s) \ge 0$, the summation term $\sum (\gamma_k - 1)\Phi(s_k)$ is non-positive. Therefore, the reward is strictly bounded by the potential difference. The maximum value occurs when $\Phi(s_0)=0$ and $\Phi(s_{K^-})=1$:
\begin{equation}
    \max_{\tau^-} R_{\text{total}}(\tau^-) \le \alpha (1 - 0) = \alpha
\end{equation}

\paragraph{2. Lower Bound for Correct Trajectories ($\tau^+$).}
For a correct trajectory, every segment contributes a base outcome reward of $1$. We seek the global minimum reward, which corresponds to the adversarial worst-case scenario:
\begin{enumerate}
    \item The trajectory length is minimal ($K^+=1$), minimizing the dense outcome contribution.
    \item The potential function is adversarial, dropping from maximum to minimum ($\Phi=1 \to 0$).
    \item The length penalty is maximized ($\gamma_k = \gamma_{\min}$).
\end{enumerate}
Substituting these conditions into the reward equation:
\begin{equation}
    \min_{\tau^+} R_{\text{total}}(\tau^+) = 1 + \alpha (\gamma_{\min} \cdot 0 - 1) = 1 - \alpha
\end{equation}

\paragraph{3. Consistency Theorem.}
To guarantee Strong Task Consistency, we require the lower bound of correct trajectories to exceed the upper bound of incorrect ones:
\begin{equation}
    \min R(\tau^+) > \max R(\tau^-) \implies 1 - \alpha > \alpha.
\end{equation}
Solving for $\alpha$, we obtain the sufficient condition:
\begin{equation}
    \alpha < 0.5
\end{equation}
This derivation proves that by setting $\alpha < 0.5$, the dense outcome signal ($R_{\text{outcome}}$) serves as a dominant anchor. Even in the presence of extreme shaping noise or length penalties, the model guarantees a strictly higher reward for correct reasoning compared to any incorrect shortcut.

\subsection{Theoretical Bound of the Discount Factor ($\gamma_{\min}$)}
\label{app:gamma_bound}

A key empirical finding from the ablation study is that setting $\gamma_{\min}=0.7$ leads to performance collapse, while $\gamma_{\min}=0.9$ works well. In this section, we provide a theoretical explanation for this phenomenon by deriving a critical lower bound on $\gamma$ grounded in the concept of \textit{Reward Sign Consistency}.

\paragraph{Reward Sign Consistency.}
We require that any positive reasoning step—one that strictly improves the solvability of the current state ($\Phi(s_{k+1}) > \Phi(s_k)$)—must receive a strictly positive shaping reward. Formally, for the shaping term $F_k = \gamma_k \cdot \Phi(s_{k+1}) - \Phi(s_k)$ to be positive, we need:
\begin{equation}
    \gamma_k \cdot \Phi(s_{k+1}) - \Phi(s_k) > 0 \implies \gamma_k > \frac{\Phi(s_k)}{\Phi(s_{k+1})}.
\end{equation}
This constraint is most stringent when transitioning from a high-potential state near the solution. In the worst case, the model improves from $\Phi(s_k) = 7/8$ to $\Phi(s_{k+1}) = 1$, yielding:
\begin{equation}
    \gamma_k > \frac{7/8}{1} = 0.875.
\end{equation}
This constitutes a critical lower bound: any $\gamma_{\min} \le 0.875$ risks sign reversal, causing the model to be penalized for making correct progress.

\paragraph{Illustrative Example.}
Table~\ref{tab:gamma_bound} concretizes this with a representative transition ($\Phi_k = 5/8 \to \Phi_{k+1} = 7/8$, a $+25\%$ improvement in solvability). As $\gamma$ decreases below $0.8$, the shaping reward flips from positive to negative, directly conflicting with the learning objective.

\begin{table}[h]
    \centering
    
    \small
    \begin{tabular}{cccc}
        \toprule
        $\gamma$ & $\gamma \cdot \Phi_{k+1}$ & $\Phi_k$ & $F$ \\
        \midrule
        1.0 & 0.875 & 0.625 & $+$0.250 \\
        0.9 & 0.788 & 0.625 & $+$0.163 \\
        0.8 & 0.700 & 0.625 & $+$0.075 \\
        \textbf{0.7} & \textbf{0.613} & \textbf{0.625} & $\mathbf{-}$\textbf{0.013} \\
        0.6 & 0.525 & 0.625 & $-$0.100 \\
        \bottomrule
    \end{tabular}
    \caption{Shaping reward $F = \gamma \cdot \Phi_{k+1} - \Phi_k$ for a positive reasoning step ($\Phi_k\!=\!5/8,\ \Phi_{k+1}\!=\!7/8$).}
    \label{tab:gamma_bound}
\end{table}

\paragraph{Conclusion.}
Our choice of $\gamma_{\min}=0.9$ maintains a safe margin above the critical threshold ($0.9 > 0.875$), ensuring efficiency is only optimized when doing so does not conflict with correctness. Setting $\gamma_{\min}=0.7 \ll 0.875$ violates Reward Sign Consistency, making training collapse mathematically inevitable.

\subsection{Derivation of Mechanism Properties}
\label{app:properties}

In this section, we provide the rigorous mathematical derivation for the two key inductive biases encoded in our advantage formulation: \textbf{Stage Awareness} and the \textbf{Token Efficiency}.

Recall the shaping term $F_k$ defined in Equation~\eqref{eq:final_advantage}:
\begin{equation}
    F_k(L_k, \Phi(s_k), \Delta_k) = \gamma_k(L_k) \cdot \Phi(s_{k+1}) - \Phi(s_k).
\end{equation}
Let $\Delta_k = \Phi(s_{k+1}) - \Phi(s_k)$ denote the semantic progress. Substituting $\Phi(s_{k+1}) = \Phi(s_k) + \Delta_k$, we rewrite $F_k$ as:
\begin{equation}
  \begin{aligned}
    F_k &= \gamma_k(L_k) \cdot (\Phi(s_k) + \Delta_k) - \Phi(s_k) \\ 
    &= \gamma_k(L_k) \Delta_k - (1 - \gamma_k(L_k)) \Phi(s_k).
    \label{eq:app_decomposition}
  \end{aligned}
\end{equation}
This decomposition highlights that the reward consists of a \textit{discounted gain} $\gamma_k \Delta_k$ minus a \textit{baseline penalty} $(1 - \gamma_k)\Phi(s_k)$.

\paragraph{Property 1: Stage Awareness.}
\textbf{Proposition.} For a fixed amount of progress $\Delta_k$ and a fixed segment length $L_k$, the shaping reward $F_k$ decreases as the baseline potential $\Phi(s_k)$ increases.

\textit{Proof.}
We analyze the sensitivity of the reward with respect to the starting state's potential $\Phi(s_k)$ by taking the partial derivative of Equation~\eqref{eq:app_decomposition}:
\begin{equation}
    \frac{\partial F_k}{\partial \Phi(s_k)} = \frac{\partial}{\partial \Phi(s_k)} \left[ \gamma_k \Delta_k - (1 - \gamma_k)\Phi(s_k) \right].
\end{equation}
Since $\gamma_k$ depends only on $L_k$ and $\Delta_k$ is fixed, they are treated as constants. The derivative simplifies to:
\begin{equation}
    \frac{\partial F_k}{\partial \Phi(s_k)} = -(1 - \gamma_k(L_k)) = \gamma_k(L_k) - 1.
\end{equation}
Given that the discount factor is bounded by $\gamma_k \in [\gamma_{\min}, 1]$, we have $\gamma_k - 1 \le 0$.
Thus:
\begin{equation}
    \frac{\partial F_k}{\partial \Phi(s_k)} \le 0
\end{equation}
\textit{Interpretation.} The derivative is strictly negative (assuming $L_k > 0$ so that $\gamma_k < 1$). This mathematically confirms that:
\begin{itemize}
    \item \textbf{Low Baseline (Low $\Phi(s_k)$):} The penalty term $(1-\gamma_k)\Phi(s_k)$ is small. The model retains most of the gain $\Delta_k$.
    \item \textbf{High Baseline (High $\Phi(s_k)$):} The penalty term is maximized. To achieve the same net reward $F_k$, the model must generate a significantly larger $\Delta_k$. This effectively suppresses potential inflation in high-confidence states.
\end{itemize}

\paragraph{Property 2: Token Efficiency.}
\textbf{Proposition.} For a fixed progress $\Delta_k$ starting from a fixed state $\Phi(s_k)$, the shaping reward $F_k$ strictly decreases as the segment length $L_k$ increases.

\textit{Proof.}
We analyze the sensitivity of the reward with respect to segment length $L_k$.
First, recall the linear definition of $\gamma_k(L_k)$ from Equation~\eqref{eq:dynamic_gamma}:
\begin{equation}
    \gamma_k(L_k) = 1 - c \cdot L_k, \quad \text{where } c = \frac{1 - \gamma_{\min}}{L_{\text{ref}}} > 0
\end{equation}
The derivative of the discount factor with respect to length is:
\begin{equation}
    \frac{d \gamma_k}{d L_k} = -c < 0
\end{equation}
Now, taking the partial derivative of $F_k$ (Equation~\eqref{eq:app_decomposition}) with respect to $L_k$:
\begin{equation}
\begin{aligned}
    \frac{\partial F_k}{\partial L_k} &= \frac{\partial}{\partial L_k} \left[ \gamma_k(L_k) (\Phi(s_k) + \Delta_k) - \Phi(s_k) \right] \\
    &= (\Phi(s_k) + \Delta_k) \cdot \frac{d \gamma_k}{d L_k} \\
    &= \Phi(s_{k+1}) \cdot (-c)
\end{aligned}
\end{equation}
Since potential $\Phi(s_{k+1}) \ge 0$ and $c > 0$, the derivative is strictly non-positive:
\begin{equation}
    \frac{\partial F_k}{\partial L_k} \le 0
\end{equation}
\textit{Interpretation.} The reward monotonically decreases as the segment becomes longer.
By examining the decomposition $F_k = \gamma_k \Delta_k - (1 - \gamma_k)\Phi(s_k)$, we see that increasing $L_k$ (which decreases $\gamma_k$) imposes a {double penalty}:
\begin{enumerate}
    \item \textbf{Diminishing Returns:} The term $\gamma_k \Delta_k$ shrinks, meaning the same semantic progress is worth less if it takes longer to generate.
    \item \textbf{Escalating Tax:} The term $(1 - \gamma_k)$ grows, increasing the penalty proportional to the current state potential.
\end{enumerate}
This creates a compounding pressure on the model to be concise, especially when the current potential $\Phi(s_k)$ is high.

\section{Experimental Details}
\label{sec:details}

\subsection{Details of Main Experiments}
\label{app:main_result}

In this section, we provide comprehensive details regarding the experimental setup, including training hyperparameters, infrastructure configurations, and evaluation protocols.

\subsubsection{Training Configuration}
\label{app:train_config}
We conduct our main experiments on three base models: DeepSeek-R1-Distill-Qwen-1.5B \citep{guo2025deepseek}, DeepScaleR-1.5B-Preview \citep{deepscaler2025}, and Qwen3-4B \citep{yang2025qwen3}.
We implement our training pipeline based on the \texttt{VeRL} framework \cite{sheng2024hybridflow}, optimizing for efficient hybrid data and model parallelism. We utilize \texttt{vLLM} for high-throughput rollout generation with a tensor parallel size of 1. To manage memory efficiency, we enable gradient checkpointing and set the optimizer/parameter offload to \texttt{True} via FSDP (Fully Sharded Data Parallel).

We employ the \textit{clip-higher} PPO variant \cite{yu2025dapo} to ensure training stability, with clipping thresholds set to $\epsilon_{\text{high}} = 0.28$ and $\epsilon_{\text{low}} = 0.2$. The KL divergence penalty coefficient is set to 0 to prioritize direct reward optimization, relying on the clipping mechanism for policy constraints.
To accommodate different model capacities, we adjust the maximum response length and the number of segments $K$:
\begin{itemize}
    \item \textbf{1.5B Models} (DeepSeek-R1-Distill-Qwen, DeepScaleR): Max response length = 8,192 tokens; Segment count $K=8$.
    \item \textbf{4B Model} (Qwen3): Max response length = 16,384 tokens; Segment count $K=16$.
\end{itemize}
A detailed summary of the training hyperparameters is provided in Table \ref{tab:train_hyperparams}.

\begin{table}[h]
    \centering

    \resizebox{0.9\linewidth}{!}{
    \begin{tabular}{l c}
        \toprule
        \textbf{Hyperparameter} & \textbf{Value} \\
        \midrule
        Global Batch Size & 128 \\
        Mini-Batch Size & 32 \\
        Gradient Accumulation Steps & 4 \\
        Learning Rate & $1 \times 10^{-6}$ \\
        Warmup Ratio & 0.1 \\
        Total Epochs & 1 \\
        Max Training Steps & 360 \\
        Clip Ratio (High / Low) & 0.28 / 0.20 \\
        KL Coefficient & 0.0 \\
        Entropy Coefficient & 0.0 \\
        Max Prompt Length & 1,024 \\
        Num Generation  & 8 \\
        Use Vllm & True \\
        BF16 & True \\
        Process Reward Coeff ($\alpha$) & 0.3 \\
        Discount Lower Bound ($\gamma_{\min}$) & 0.9 \\
        \bottomrule
    \end{tabular}
        }
    \caption{Detailed training hyperparameters.}
    \label{tab:train_hyperparams}

\end{table}

\subsubsection{Evaluation Protocol}
\label{app:eval_config}

\paragraph{Inference Settings.}

We perform zero-shot evaluation across five benchmarks: AIME 2024 \cite{maa2024aime}, AIME 2025 \cite{maa2025aime}, AMC 2023 \cite{maa2023amc}, MinervaMATH \cite{lewkowycz2022solving}, and MATH500 \cite{hendrycks2021measuring}. To balance exploration and stability, we use a temperature of $T=0.6$, top-$p=0.95$, and top-$k=40$. To prevent truncation of long reasoning chains, the maximum generation length is set to 32,768 tokens. Answer extraction and correctness verification are performed using the \texttt{Math-Verify} \footnote{https://github.com/huggingface/Math-Verify} library, ensuring robust parsing of mathematical expressions.

\paragraph{Metric: Avg@N Pass@1.}
Given the high variance inherent in reasoning model outputs, reporting a single pass rate from a small sample size can be unstable. To ensure statistical reliability, we report the {avg@N Pass@1} metric. Specifically, for each problem, we sample $N$ independent responses and compute the average accuracy. The value of $N$ is adaptively scaled based on the dataset size to maintain a sufficiently large total sample pool (approx. 1,000 samples per benchmark) for low-variance estimation.
Table \ref{tab:eval_datasets} details the sampling configuration for each benchmark.

\begin{table}[h]
    \centering

    \resizebox{\linewidth}{!}{
    \begin{tabular}{l c c c}
        \toprule
        \textbf{Benchmark} & \textbf{\# Probs} & \textbf{Samples ($N$)} & \textbf{Total Resps.} \\
        \midrule
        AIME 2024 & 30 & 32 & 960 \\
        AIME 2025 & 30 & 32 & 960 \\
        AMC 2023 & 40 & 32 & 1,280 \\
        MinervaMATH & 272 & 4 & 1,088 \\
        MATH-500 & 500 & 2 & 1,000 \\
        \bottomrule
    \end{tabular}
        }
    \caption{Evaluation dataset specifications and sampling configurations.}
    \label{tab:eval_datasets}

\end{table}

\subsection{Details of Sensitivity Analysis of Potential Gains}
\label{app:sensitivity_details}

In \S \ref{sec:sensitivity_analysis}, we verify the core motivation of SHAPE by analyzing the correlation between intermediate potential gains and final outcomes. Here we describe the data collection process, grouping strategy, and the centered regression methodology used to generate Figure \ref{fig:sensitivity}.

\paragraph{Data Collection and Sampling.}
The analysis is performed on the intermediate potential estimates generated during the SHAPE training process. Unlike standard inference trajectories, these data points represent the on-policy potential estimations calculated at each segment boundary. Specifically, for a training batch with $B$ prompts, each rollout is divided into $K$ segments. At the boundary of segment $s_k$, the model executes branch rollouts to estimate the state potential $\Phi(s_k)$.

To analyze the relationship between potential evolution and final success, we recorded the full trajectory of potential transitions $(\Phi(s_k) \rightarrow \Phi_{k+1})$ and the corresponding final binary outcome $y \in \{0, 1\}$. The training process spans 360 steps (batches). To ensure computational efficiency while maintaining temporal diversity across different training stages, we applied a strided sampling strategy with an interval of 10 steps. This resulted in a total of 36 batches of data, comprising approximately $4.1 \times 10^5$ segment transitions for analysis.

\paragraph{Grouping and Filtering.}
We categorize the transitions based on their starting potential $\Phi(s_k)$:
\begin{itemize}
    \item \textbf{Low Start Regime:} $\Phi(s_k) \le 0.25$. This corresponds to states where the reasoning path is potentially flawed or confused (e.g., $0/8$ to $2/8$ correctness).
    \item \textbf{High Start Regime:} $\Phi(s_k) \ge 0.5$. This corresponds to states where the reasoning path is already partially or mostly correct.
\end{itemize}
To ensure a robust linear fit, we apply a specific filter to the Low Start group. We exclude floor effect transitions where the potential drops significantly to zero (specifically, $Gain \le -0.24$ when starting from $\Phi(s_k) \le 0.25$), as these boundary conditions represent saturation points that can skew the linearity of the sensitivity estimation.

\paragraph{Centered Regression Methodology.}
We employ Ordinary Least Squares (OLS) regression on the filtered dataset to determine the slope $m$ (sensitivity) for each group. However, since High Start states naturally possess higher baseline success rates than Low Start states, a direct overlay of their regression lines results in parallel lines with distinct intercepts, obscuring the comparison of their slopes.

To resolve this and visually highlight the \textit{marginal effect}, we employ a {Centered Regression} approach. For each group $g \in \{\text{Low}, \text{High}\}$, we first calculate the raw intercept $\beta_0^{(g)}$ from the dataset. We then define the centered outcome $y'$ as:
\begin{equation}
    y' = y - \beta_0^{(g)}.
\end{equation}
This transformation aligns the y-intercept of the regression lines to zero, effectively isolating the relative change in success probability attributable to the potential gain $\Delta$. The slopes reported in the main text ($m_{low} \approx 0.65, m_{high} \approx 0.55$) are statistically significant and confirm the higher marginal utility of gains in the Low Start regime.

\subsection{Details of Potential Gain Analysis}
\label{app:value_gain_analysis}

A core premise of SHAPE is that not all potential gains are equal. While MRT treats a potential improvement from 0.1 to 0.2 equally to an improvement from 0.8 to 0.9, we argue that gains from a \textit{low starting potential} are more critical, as they represent early error correction and path rectification. To verify whether SHAPE successfully incentivizes this behavior, we analyze the distribution of potential gain contributions across different starting states.

To rigorously quantify the evolution of the model's reasoning strategy (\S \ref{sec:reasoning_strategy}), we implemented a statistical analysis of step-level potential trajectories. The detailed procedure is as follows:

\paragraph{Data Collection.}
We recorded the step-level potential trajectories for every training batch. As defined in Equation~\eqref{eq:prelim_mrt_phi}, the potential at segment boundary $s_k$, denoted as $\Phi(s_k)$, is estimated via the rollout success rate. For a rollout width of $N=8$, $\Phi(s_k)$ is calculated as $n_k/N$, where $n_k$ is the number of correct outcomes. Consequently, the discrete set of possible potential values is $\{0, 0.125, \dots, 1.0\}$. We analyzed three datasets: the first 10 steps of MRT training (Initial), the last 10 steps of MRT training, and the last 10 steps of SHAPE training.

\paragraph{Gain Calculation and Binning.}
We calculate the potential gain for each segment transition as $\Delta_k = \Phi(s_{k+1}) - \Phi(s_k)$. These gains are then aggregated based on their starting potential $\Phi(s_k)$. We group the 8 possible starting states (excluding the terminal state $1.0$ which cannot yield positive gain) into four categories:
\begin{itemize}
    \item \textbf{Low Start:} $\Phi(s_k) \in \{0/8, 1/8\}$
    \item \textbf{Mid-Low Start:} $\Phi(s_k) \in \{2/8, 3/8\}$
    \item \textbf{Mid-High Start:} $\Phi(s_k) \in \{4/8, 5/8\}$
    \item \textbf{High Start:} $\Phi(s_k) \in \{6/8, 7/8\}$
\end{itemize}

\paragraph{Normalization.}
Since raw potential gains can be negative (indicating potential degradation), comparing absolute sums across different training stages is challenging. To visualize the relative contribution distribution, we apply a global normalization:
1. We compute the average raw gain for each category across all samples.
2. We identify the global minimum average gain $G_{\min}$ across all methods and categories.
3. We apply a floor shift to ensure non-negativity: $G'_{\text{cat}} = G_{\text{cat}} - (G_{\min} - \epsilon)$, where $\epsilon=0.03$.
4. Finally, we calculate the percentage contribution of each category relative to the total shifted gain of the method: $P_{\text{cat}} = \frac{G'_{\text{cat}}}{\sum G'} \times 100\%$.

This metric effectively highlights which state capability (repairing low-potential states vs. improving high-potential states) contributes most to the model's learning progress.

\subsection{Details of Adaptive Computation Analysis}
\label{app:adaptive_scaling}

In \S \ref{sec:adaptive_computation}, we presented the length-difficulty alignment analysis. Here, we detail the dataset construction, difficulty estimation, and visualization standards.

\paragraph{Dataset and Difficulty Estimation.}
To ensure a representative analysis, we constructed a specific {Difficulty Calibration Subset.} We randomly sampled 20 problems from each of the 5 evaluation benchmarks (AIME 2024, AIME 2025, AMC 2023, MinervaMATH, and MATH500), resulting in a total of 100 distinct problems.

To establish an objective difficulty score ($D$), we utilized the base model (\textit{DeepSeek-R1-Distill-Qwen-1.5B}) to avoid bias from post-training. For each problem in this subset, we generated $N=10$ independent responses. The difficulty is defined as:
\begin{equation}
    D = 1 - \text{PassRate}_{\text{Base}}.
\end{equation}
Based on these scores derived from the 1,000 aggregated responses, the problems are categorized into the five difficulty bins defined in the main text.

\paragraph{Evaluation Procedure.}
After binning, we evaluated the trained {GRPO} and {SHAPE} models on this specific subset. We recorded the token lengths of their generated responses to analyze how each model adapts its reasoning depth relative to the pre-determined difficulty levels.

\paragraph{Box Plot Interpretation.}
To aid in interpreting Figure \ref{fig:adaptive_scaling}, the components of the box plots are defined as follows:
\begin{itemize}
    \item \textbf{Central Line:} Represents the \textit{median} response length of the model in that difficulty bin.
    \item \textbf{Box Limits \& Height:} The box spans from the 25th percentile ($Q1$) to the 75th percentile ($Q3$), known as the \textit{Interquartile Range (IQR)}. The height of the box reflects the variance or spread of the middle 50\% of the data; a shorter box indicates more consistent reasoning length.
    \item \textbf{Whiskers:} The vertical lines extending above and below the box indicate the range of the data ($1.5 \times \text{IQR}$), excluding outliers.
\end{itemize}

\section{Out-of-Distribution Generalization}
\label{app:ood}

Since SHAPE updates model parameters with strong inductive biases (e.g., conciseness), a natural concern is whether training on mathematical tasks causes the model to overfit, degrading performance on out-of-distribution (OOD) domains. To address this, we evaluate our math-trained models on two challenging OOD benchmarks:

\begin{itemize}
    \item \textbf{GPQA Diamond} \citep{rein2024gpqa}: A rigorous dataset of graduate-level questions in physics, chemistry, and biology (198 questions from the Diamond subset).
    \item \textbf{LiveCodeBench (V5)} \citep{jain2024livecodebench}: A contamination-free benchmark of competitive programming problems from LeetCode, AtCoder, and Codeforces, updated continuously to prevent data leakage.
\end{itemize}

Results are reported in Table~\ref{tab:ood_results} as zero-shot Pass@1 accuracy (\%).

\begin{table}[h]
    \centering
    \small
    \setlength{\tabcolsep}{4pt}
    \begin{tabular}{l cc}
        \toprule
        \textbf{Method} & \textbf{GPQA} & \textbf{LiveCodeBench} \\
        \midrule
        \rowcolor{ModelGray}
        \multicolumn{3}{l}{\textbf{\textit{DS-Qwen-1.5B}}} \\
        GRPO  & 36.6 & 19.1 \\
        MRT   & 35.9 & 18.4 \\
        \rowcolor{HighlightRow}
        SHAPE & \textbf{38.4} & \textbf{22.3} \\
        \midrule
        \rowcolor{ModelGray}
        \multicolumn{3}{l}{\textbf{\textit{DeepScaleR-1.5B}}} \\
        GRPO  & 40.6 & 23.4 \\
        MRT   & \textbf{41.9} & 25.2 \\
        \rowcolor{HighlightRow}
        SHAPE & 41.7 & \textbf{25.8} \\
        \midrule
        \rowcolor{ModelGray}
        \multicolumn{3}{l}{\textbf{\textit{Qwen3-4B}}} \\
        GRPO  & 52.8 & 54.4 \\
        MRT   & 52.5 & 54.1 \\
        \rowcolor{HighlightRow}
        SHAPE & \textbf{54.4} & \textbf{56.7} \\
        \bottomrule
    \end{tabular}
    \caption{Zero-shot performance on OOD benchmarks.}
    \label{tab:ood_results}
\end{table}

SHAPE consistently matches or outperforms the baselines across both domains. Two findings are noteworthy. First, math-focused SHAPE training yields positive transfer to coding (e.g., $+3.2\%$ on LiveCodeBench for the 1.5B model), consistent with recent observations in AceMath-RL \citep{liu2025acemath} that logical rigor learned from mathematical reasoning transfers effectively to algorithmic problem-solving. Second, on GPQA, SHAPE maintains or improves over GRPO, indicating that the inductive bias toward concise reasoning does not degrade general knowledge application. Together, these results confirm that SHAPE enhances the underlying efficiency and logical consistency of the policy—transferable virtues that extend well beyond the mathematical domain.

\section{Use of AI Assistants and Artifact Usage}
\label{app:ethics}

\paragraph{Use of AI Assistants.}
In accordance with the ACL policy on AI assistance, we acknowledge the use of AI assistants solely for the purpose of linguistic polishing and grammatical error correction. The core scientific ideas, experimental designs, and data analyses were conducted entirely by the human authors. No text generated by the AI tool contains novel scientific claims or results.

\paragraph{Artifact Licenses and Intended Use.}
We utilize publicly available models and datasets to conduct our experiments.
\begin{itemize}
    \item \textbf{Models:} We employ open-weights models including \textit{DeepSeek-R1-Distill-Qwen-1.5B}, \textit{Qwen3-4B}, and \textit{DeepScaleR-1.5B-Preview}. These models are used in accordance with their respective open-source licenses and intended use for research purposes.
    \item \textbf{Datasets:} We evaluate on standard mathematical benchmarks including AIME, AMC, MATH, and MinervaMATH, as well as out-of-distribution benchmarks including GPQA Diamond and LiveCodeBench. These datasets are widely distributed for academic research. We adhere to their respective terms of use, utilizing them strictly for non-commercial research evaluation.
\end{itemize}
We confirm that our use of these artifacts is consistent with their intended scientific use cases.

\end{document}